\documentclass[lettersize,journal]{IEEEtran}
\usepackage{amsmath,amsfonts}
\usepackage{algorithmic}
\usepackage{algorithm}
\usepackage{array}
\usepackage[caption=false,font=normalsize,labelfont=sf,textfont=sf]{subfig}
\usepackage{textcomp}
\usepackage{stfloats}
\usepackage{url}
\usepackage{verbatim}
\usepackage{graphicx}
\usepackage{cite}

\usepackage{graphicx}  
\usepackage{afterpage} 
\usepackage{caption}   
\usepackage{cuted}
\usepackage{algorithm}
\usepackage{algorithmic}

\usepackage{amsmath}

\usepackage{subfig}

\usepackage{xcolor}

\usepackage{tabularray}
\usepackage{colortbl}
\usepackage{multirow}
\usepackage{hhline}

\usepackage{xcolor}
\definecolor{red}{rgb}{1,0.6,0.6}
\definecolor{orange}{rgb}{1,0.8,0.6}
\definecolor{yellow}{rgb}{1,0.98,0.6}

\begin{document}
\renewcommand{\arraystretch}{1.2} 
\setlength{\tabcolsep}{4pt} 

\title{VINGS-Mono: Visual-Inertial Gaussian Splatting Monocular SLAM in Large Scenes}

\author{Ke Wu, Zicheng Zhang, Muer Tie, Ziqing Ai, Zhongxue Gan, Wenchao Ding
         \thanks{Academy for Engineering and Technology, Fudan University, Shanghai, China. E-Mail: dingwenchao@fudan.edu.cn, kewu23@m.fudan.edu.cn}
}



\maketitle
\begin{strip}
    \centering
    \vspace{-1em}
    \includegraphics[width=\textwidth]{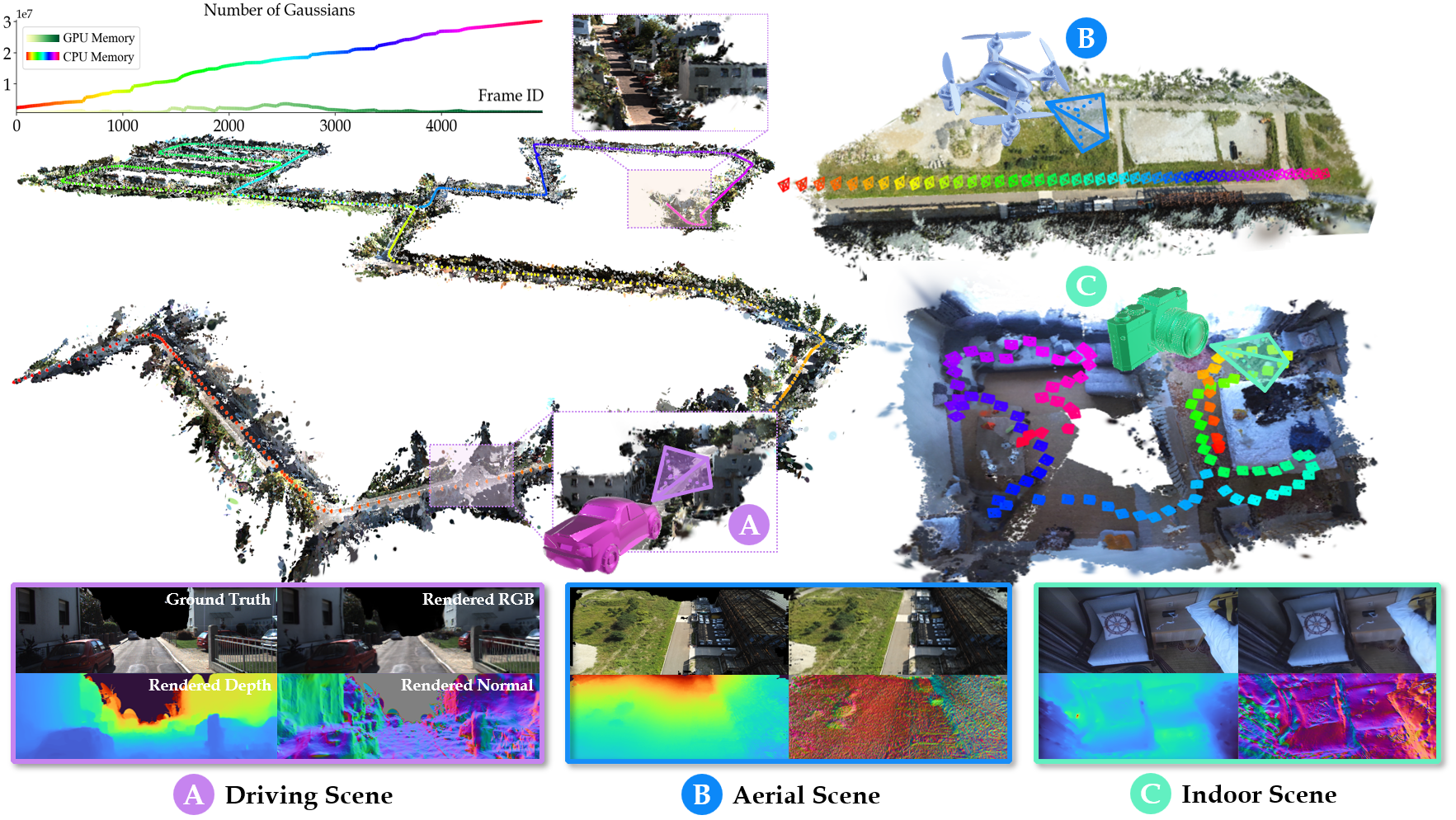} 
    \captionof{figure}{\textbf{VINGS-Mono's estimated trajectory and reconstructed gaussian map of three different scenes.} 
Our method effectively estimates poses and reconstructs high-quality Gaussian maps across large-scale driving scenarios, aerial drone views, and indoor environments. Particularly for the \textbf{driving scene} on the left, the trajectory spans 3.7 kilometers and includes a Gaussian map containing 32.5 million Gaussian ellipsoids. During training, we track the number of Gaussians and zoom in on specific areas to improve visualization clarity. (Project page: https://vings-mono.github.io)}
    \label{cover_image}
    \vspace{1em} 
\end{strip}


\begin{abstract}
VINGS-Mono is a monocular (inertial) Gaussian Splatting (GS) SLAM framework designed for large scenes. The framework comprises four main components: VIO Front End, 2D Gaussian Map, NVS Loop Closure, and Dynamic Eraser. In the VIO Front End, RGB frames are processed through dense bundle adjustment and uncertainty estimation to extract scene geometry and poses. Based on this output, the mapping module incrementally constructs and maintains a 2D Gaussian map. Key components of the 2D Gaussian Map include a Sample-based Rasterizer, Score Manager, and Pose Refinement, which collectively improve mapping speed and localization accuracy. This enables the SLAM system to handle large-scale urban environments with up to 50 million Gaussian ellipsoids. To ensure global consistency in large-scale scenes, we design a Loop Closure module, which innovatively leverages the Novel View Synthesis (NVS) capabilities of Gaussian Splatting for loop closure detection and correction of the Gaussian map. Additionally, we propose a Dynamic Eraser to address the inevitable presence of dynamic objects in real-world outdoor scenes. Extensive evaluations in indoor and outdoor environments demonstrate that our approach achieves localization performance on par with Visual-Inertial Odometry while surpassing recent GS/NeRF SLAM methods. It also significantly outperforms all existing methods in terms of mapping and rendering quality. Furthermore, we developed a mobile app and verified that our framework can generate high-quality Gaussian maps in real time using only a smartphone camera and a low-frequency IMU sensor. To the best of our knowledge, VINGS-Mono is the first monocular Gaussian SLAM method capable of operating in outdoor environments and supporting kilometer-scale large scenes.

\end{abstract}



\section{Introduction}

\IEEEPARstart{A}{n} information-rich, geometrically dense map is essential for a robot's environmental perception and scene understanding. 3D Gaussian Splatting (3DGS)~\cite{3DGS}  has rapidly gained popularity due to its exceptional rendering speed and high-quality visuals. 3DGS enhances SLAM systems by providing detailed scene information and enabling novel view synthesis. Furthermore, due to Gaussian Splatting's differentiable rendering process, we can construct the dense maps using only low-cost RGB supervision.

Existing 3DGS SLAM systems~\cite{SplaTAM,rtgslam} primarily focus on a limited number of displayed objects or small indoor spaces, using depth cameras as input and leveraging traditional SLAM front end or depth point cloud ICP for localization and Gaussian updates. Outdoor GS-SLAM methods~\cite{livgaussmap} are scarce and restricted to reconstructing scenes within a few hundred meters, relying heavily on high-beam LiDAR sensors. However, depth cameras perform poorly in outdoor settings, and the high cost of LiDAR sensors has limited their adoption in consumer applications. Given constraints in size, weight, and power, a low-cost camera paired with an IMU forms the minimal sensor suite for SLAM implementation. Therefore, developing a robust monocular (inertial) GS-SLAM system capable of handling large-scale environments is both essential and urgent.

Currently, monocular input-supported 3DGS SLAM systems~\cite{MONO-GS, hhuang2024photoslam}, which initialize Gaussians using random or sparse feature points, are unable to handle large-scale, fast-moving scenes due to their vulnerability to pose drift and geometry noise. Furthermore, significant accumulated errors are commonly observed in large-scale environments. These errors are typically mitigated through loop closure. Traditional loop closure methods, relying on descriptors or network feature vectors, require additional encoding and storage of bag-of-words models, which is inefficient and leads to performance degradation as the scene scale increases. GO-SLAM~\cite{GO-SLAM}, on the other hand, identifies loop closures by maintaining the co-visibility matrix between frames, but this results in quadratic storage demands and increased computational overhead.

Developing an efficient and high-fidelity monocular GS-SLAM for large-scale scenes faces several significant challenges. First, representing large, street-level scenes requires managing tens of millions of Gaussians, which is both storage-intensive and computationally demanding. Second, monocular setups suffer from severe scale drift, which undermines the accuracy and reliability of the reconstructed scenes.  Furthermore, significant cumulative errors arise in large-scale environments. While traditional loop closure techniques are effective at optimizing landmark-based maps, correcting a dense Gaussian map after detecting a loop closure is highly challenging and often requires retraining on all historical frames. Lastly, the presence of dynamic objects in large urban environments poses significant challenges, as they generate considerable artifacts and noise in the Gaussian map, further complicating the optimization process.

In this paper, we introduce VINGS-Mono, a monocular (inertial) Gaussian Splatting SLAM framework that supports large-scale urban scenes. The framework consists of four main modules: VIO Front End, 2D Gaussian Map, NVS Loop Closure, and Dynamic Object Eraser. To address challenges in Gaussian map storage and optimization efficiency, we develop a score manager to manage the 2D Gaussian Map by integrating local and global map representations. Additionally, we design a sample rasterizer to accelerate the backpropagation algorithm of Gaussian Splatting, significantly improving its computational efficiency. To enhance tracking accuracy and mitigate the inevitable drift encountered in large-scale scenarios, we propose a single-to-multi pose refinement module. This module back-propagates rendering errors from a single frame to optimize the poses of all frames within the frustum's field of view, improving overall pose consistency. For accumulated errors, we utilize the novel view synthesis (NVS) capability of Gaussian Splatting for loop closure detection. We further propose an efficient loop correction method capable of simultaneously adjusting millions of Gaussian attributes upon detecting a loop. Finally, to address the impact of dynamic objects on mapping, we design a heuristic semantic segmentation mask generation method based on re-rendering loss. This method ensures that dynamic objects are effectively handled, enhancing the robustness of the mapping process.

Our contributions can be summarized as follows:
\begin{itemize} 
    \item We are the first monocular (inertial) GS-based SLAM system capable of operating in outdoors and support kilometer-scale urban scenes.
    \item We propose a 2D Gaussian Map module, including a sample rasterizer, score manager, and single-to-multi pose refinement, ensuring that our method could achieve accurate localization and build high-quality gaussian maps in real time.  
    \item We introduce a GS-based loop detection method, along with an efficient approach that can correct tens of millions of Gaussian attributes in a single operation upon loop detection, effectively eliminating accumulated errors and ensuring the global consistency of the map.  
    \item Comprehensive experiments on different scenes (indoor environments, aerial drone view and driving scenes) demonstrate that VINGS-Mono outperforms existing approaches in both rendering and localization performance. Furthermore, we developed a mobile app and carried out real-world experiments to demonstrate the practical reliability of our method.  
\end{itemize}

\section{Related Works}

In this section, we review related works in Gaussian Splatting SLAM, Visual Loop Closure, and Large-Scale Visual SLAM, as they are highly relevant to our framework. These topics cover the core aspects of our method: scene representation with Gaussian splatting, ensuring global consistency through loop closure, and addressing scale drift and memory consumption in large-scale environments.

\subsection{Gaussian Splatting SLAM}
3D Gaussian Splatting, with its differentiable nature and fast rendering speed, has emerged as a promising scene representation in SLAM systems. Compared with traditional explicit map representation such as voxel grids\cite{KinectFusion}, surfels\cite{ElasticFusion}, pointclouds\cite{VINS-MONO, ORBSLAM3_TRO}, GS representation provides a denser map, the ability to supervise geometry through differentiable color rasterization, and novel view synthesis capabilities.

The initial GS-SLAM method SplaTAM~\cite{SplaTAM} focused on utilizing depth cameras to initialize Gaussian ellipsoids and optimized camera poses by backpropagating photometric errors from rendering loss to Gaussian positions and then to camera poses. However, this approach was highly sensitive to depth camera noise and demonstrated limited robustness. PhotoSLAM~\cite{hhuang2024photoslam} and MonoGS~\cite{MONO-GS} extended GS-SLAM to monocular settings for small indoor scenes, adding Gaussian ellipsoids through ORB feature points or random initialization and directly estimating poses with ORB-SLAM3~\cite{ORBSLAM3_TRO}. Despite their effectiveness in small-scale environments, these methods showed significant limitations in handling larger or dynamic scenarios, causing severe floaters that greatly compromised map quality in the SLAM system. Gaussian-SLAM~\cite{yugay2023gaussian} introduced innovations such as differential depth rendering and frame-to-model alignment, which significantly enhanced its overall performance. However, it exhibited inefficiencies in frame processing speed and memory usage. GS-ICP-SLAM~\cite{ha2024rgbdgsicpslam}, on the other hand, achieved notable improvements in frame processing rates by employing point cloud matching. More recently, MGS~\cite{zhu2024mgs} was proposed for monocular settings, leveraging a Multi-View Stereo Network from DPVO~\cite{teed2024deep} as a depth prior. While promising, this method faces challenges in scaling to large-scale scenes, limiting its applicability in extensive environments. LIVGaussMap~\cite{livgaussmap} introduced GS-Mapping to outdoor scenes at a scale of hundreds of meters. However, it relies on high-beam LiDAR, which is typically difficult to obtain for consumer-grade applications.

Although GS-SLAM has developed rapidly, the visual GS-SLAM systems mentioned above primarily focus on small-scale indoor scenes using datasets like TUM-RGBD~\cite{TUM_RGBD} and Replica~\cite{replica_dataset}. This significantly limits the applicability of GS-SLAM in larger-scale scenarios. To address this, we designed and developed a robust and efficient monocular (inertial) GS-SLAM method, which was tailored to the challenges of large-scale environments.

\subsection{Visual Loop Closure}

Visual Loop Closure consists of two main components, Loop Detection and Loop Correction. In large-scale environments, loop closure is essential, especially in monocular settings that lack scale information, where each segment of the trajectory has a different scale. Detecting and correcting loops enables visual SLAM systems to effectively eliminate cumulative errors and build globally consistent maps. 

In terms of loop detection, early methods used hand-crafted features such as Gist~\cite{Gist}, BRIEF~\cite{BRIEF}, and HOG~\cite{HOG} to capture the general appearance of an image through single vectors or histograms. However, these global features lacked robustness to rotation and scale changes. The advent of local descriptors like SIFT~\cite{SIFT}, ORB~\cite{ORB}, and CenSurE~\cite{Censure} enhanced feature robustness, employing visual bag of words (BoW)~\cite{Sivic2003VideoGA} or vocabulary trees~\cite{Nistr2006ScalableRW} for efficient descriptor management across frames. Despite their effectiveness, hand-crafted features struggled in dynamic conditions with variable lighting or seasonal changes. The shift to deep learning introduced adaptive feature extraction, significantly boosting the reliability of loop closure detection. Pioneering this approach, Chen et al.~\cite{CNNloopclosure} utilized features from all layers of trained networks for enhanced location recognition. Methods like NetVLAD~\cite{NetVLAD} then improved image descriptor resilience by integrating multiple features, while LoopSplat~\cite{Zhu2024LoopSplatLC} and hloc~\cite{hloc} further refined this approach. Additionally, advancements in visual foundation models have led to techniques like SALAD~\cite{salad} using DINOv2~\cite{dinov2}, significantly enhancing descriptor quality for loop detection in complex environments.

In terms of loop correction, ORB-SLAM3~\cite{ORBSLAM3_TRO} performs a map merging operation after detecting a loop. This process mainly consists of four steps, welding window assembly, merging maps, welding BA, and essential-graph optimization. These steps collectively optimize both the poses and the landmark map.  VINS-Mono~\cite{VINS-MONO} applies an extra pose graph optimization step to guarantee that the past poses are arranged in a globally consistent manner. However, for dense maps (e.g., NeRF~\cite{NeRF}, 3DGS~\cite{3DGS}), correcting the map after detecting a loop is challenging, as these maps are typically generated through training based on given poses and image pairs. Retraining them would be extremely time-consuming. GO-SLAM~\cite{GO-SLAM} attempts to correct poses and inverse depths by iteratively optimizing loop edges added to existing local keyframes and subsequently optimizing the dense map through training. However, this approach struggles to perform loop correction for large-scale drift in extensive scenes. LoopySLAM~\cite{loopyslam} and LoopSplat~\cite{Zhu2024LoopSplatLC} address scene map correction by constructing submaps, but this method also fails to resolve scale drift within the submap itself, making it difficult to adapt to monocular settings.

To address loop detection and correction in large-scale monocular settings and to explore the potential of 3DGS in place recognition, our approach innovatively utilizes the novel view synthesis capabilities of Gaussian Splatting, allowing us to perform loop detection using only the gaussian map. Moreover, we provide an efficient method that can correct millions of Gaussian attributes in one go upon detecting a loop, enabling us to construct high-quality, globally consistent Gaussian maps.

\subsection{Large-Scale SLAM}

In this subsection, we focus on visual-based SLAM methods tailored for large-scale environments. It is worth emphasizing that in extensive outdoor, street-level scenarios, visual SLAM is especially susceptible to scale drift and cumulative errors, posing significant challenges.

LSD-SLAM~\cite{engel2014lsd} is a pioneer in large-scale SLAM, building globally consistent maps by directly optimizing geometry on image intensities and explicitly modeling scale drift. VINS-Mono~\cite{VINS-MONO} and ORB-SLAM3~\cite{ORBSLAM3_TRO} incorporate IMU data to obtain weak scale information, ensuring localization accuracy. SelectiveVIO~\cite{selectiveVIO} and iSLAM~\cite{iSLAM} further extend this approach by utilizing neural networks to fuse visual and inertial sensor readings. However, these methods primarily focus on localization, resulting in very sparse map reconstructions. NEWTON~\cite{matsuki2024newton} successfully integrates NeRF into SLAM tasks for large-scale indoor environments (e.g., at the scale of a single floor) by constructing view-centric submaps. However, this method struggles to adapt to outdoor scenarios with fast ego-motion. LIVGaussMap~\cite{livgaussmap} and MMGaussian~\cite{mmgaussian}, by using LiDAR point clouds for initialization, extend 3D Gaussian Representation (3DGS) to outdoor SLAM tasks. Nonetheless, these methods rely heavily on LiDAR and are limited to scene scales in the range of hundreds of meters due to the large number of Gaussians involved.

We incorporate dense visual factors and IMU factors into the factor graph for optimization and design a score manager for the 2D Gaussian Map, which includes status control, storage control, and GPU-CPU transfer. This enables the reconstruction of tens of millions of Gaussian ellipsoids across kilometer-scale scenes.

\section{System Overview}

\begin{figure*}[!t]
\centering
\includegraphics[width=\textwidth]{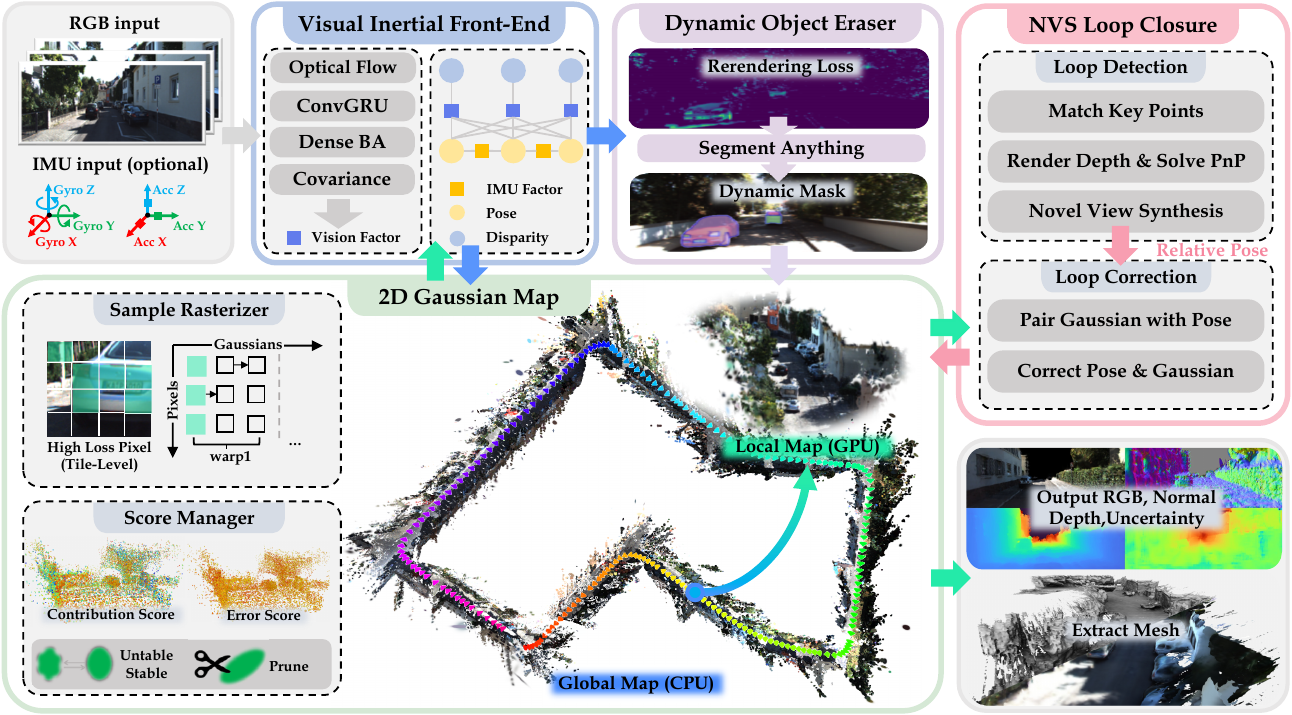}
\caption{\textbf{Pipeline of VINGS-Mono.} RGB and IMU readings are processed by the Visual Inertial Frontend to calculate pose and inverse depth. Based on this, the 2D GS Map is incrementally updated, comprising a score manager, sample rasterization, and pose refinement. The NVS Loop Closure employs novel view synthesis for efficient loop detection and correction seamlessly. Furthermore, the Dynamic Object Eraser helps minimize the impact of moving objects on the framework.}
\label{Fig:pipeline}
\end{figure*}

The pipeline of our framework is illustrated in Fig.~\ref{Fig:pipeline}. Given a sequence of RGB images and IMU readings, we first utilize the Visual Inertial Front End (Sec.~\ref{visual-inertial frontend}) to select keyframes and calculate the initial depth and pose information of the keyframes through dense bundle adjustment. Additionally, we compute the depth map uncertainty based on the covariance from the depth estimation process, filtering out geometrically inaccurate regions and sky areas. The 2D Gaussian Map module (Sec.~\ref{2d gaussian map}) incrementally adds and maintains Gaussian ellipsoids using the outputs of the visual front end. We designed a management mechanism based on contribution scores and error scores to effectively prune Gaussians. Furthermore, we propose a novel method to optimize multi-frame poses using single-frame rendering loss. To ensure scalability to large-scale urban scenes, we implemented a CPU-GPU memory transfer mechanism. In the NVS Loop Closure Module (Sec.~\ref{nvs loop closure}), we leverage the novel view synthesis capability of GS to design an innovative loop closure detection method and correct the Gaussian map through Gaussian-pose pair matching. Additionally, we integrate a Dynamic Object Eraser module (Sec.~\ref{dynamic object eraser}) that masks out transient objects like vehicles and pedestrians, ensuring consistent and accurate mapping under static scene assumptions.

\section{Visual Inertial Front End}\label{visual-inertial frontend}

The input of our Visual Inertial Front End consists of RGB images $\{I_t\}$ and IMU's acceleration and gyroscope (optional). We extract relevant features of adjacent RGB frames through the correlation volume from RAFT~\cite{raft} and feed them into the DBA module proposed in DROID-SLAM~\cite{droid-slam} to estimate inverse depths and poses (Sec.~\ref{subsec:DBA_subsection}). To enable fusion of visual data with IMU information, we use a graph optimization approach~\cite{dba-fusion, gtsam} (Sec.~\ref{subsec:vi_graph}). To prevent floaters in the map, we calculate the depth map uncertainty based on the information matrix~\cite{pv-fusion} (Sec.~\ref{subsec:depth_uncert}).

\subsection{Dense BA \& Vision Factor}\label{subsec:DBA_subsection}
The visual constraints are modeled as a Dense Bundle Adjustment (DBA) optimization problem over inverse depth and pose, where the inverse depth $\mathbf{d}_i^{-1}$ is projected ($\Pi_C^{-1}$) through pose $\mathbf{T}_{ij}$ onto frame $j$ to estimate the optical flow $\mathbf{u}_{ij}$:

\begin{equation}
\mathbf{u}_{ij} = \Pi_C (\mathbf{T}_{ij} \circ \Pi_C^{-1}(\mathbf{u}_i, \mathbf{d}_i^{-1})).
\end{equation}

For adjacent RGB frames, we construct a correlation volume using their encoded feature maps. By applying a lookup operator on  $\mathbf{u}_{ij}$, we obtain a correlation feature map. Using a GRU-based structure, we take the current image encoding, the correlation feature map, and the GRU's own hidden state as inputs, which then outputs the optical flow residual $r_{ij}$ and the weight $\mathbf{w}_{ij}$ for subsequent upsampling. To ensure real-time processing, we downsample by a factor of eight, So the resolution of the inverse depth $\mathbf{d}_i^{-1}$ is one-eighth of the original RGB image.

The GRU outputs a revision flow field $\mathbf{u}_{ij}$, and we denote the corrected correspondence as $\mathbf{u}_{ij}^* = \mathbf{r}_{ij} + \mathbf{u}_{ij}$. We can then define this DBA problem as follows, iteratively optimizing $\mathbf{d}^{-1}$ and $\mathbf{T}_{i}, \mathbf{T}_{j}$:

\begin{equation}
\mathbf{E}(\mathbf{T}, \mathbf{d}^{-1}) = \sum_{(i,j) \in \epsilon } ||\mathbf{u}_{ij}^* - \Pi_C (\mathbf{T}_{ij} \circ \Pi_C^{-1}(\mathbf{u}_i, \mathbf{d}_i^{-1}))||^2_{\Sigma_{ij}}.
\end{equation}

where $\Sigma_{ij}$ represents the diagonal matrix of $\mathbf{w}_{ij}$.

Considering a bundle of edges anchored on frame $i$ and projected to $N$ co-visible frames, the combined Hessian is constructed by positionally stacking and summing the blocks as in Eq.~\ref{hessian matrix}:

\begin{equation}
\begin{aligned}
\label{hessian matrix}
\left[\begin{array}{c}
\Sigma \mathbf{v}_{i i} \\
\mathbf{v}_{i 1} \\
\vdots \\
\mathbf{v}_{i N} \\
 \Sigma \mathbf{z}_{i i}
\end{array}\right] = \left[\begin{array}{ccccc}
\Sigma  \mathbf{B}_{i i} &  \mathbf{B}_{i 1} & \cdots &  \mathbf{B}_{i N} & \Sigma  \mathbf{E}_{i i} \\
 \mathbf{B}_{i 1}^{\top} &  \mathbf{B}_{11} & & &  \mathbf{E}_{i 1} \\
\vdots & & \ddots & & \vdots \\
 \mathbf{B}_{i N}^{\top} & & &  \mathbf{B}_{N N} &  \mathbf{E}_{i N} \\
 \Sigma  \mathbf{E}_{i i}^{\top} &  \mathbf{E}_{i 1}^{\top} & \cdots &  \mathbf{E}_{i N}^T & \Sigma  \mathbf{C}_{i i}
\end{array}\right]\left[\begin{array}{c}
\Delta  \mathbf{\xi}_{i} \\
\Delta  \mathbf{\xi}_{1} \\
\vdots \\
\Delta  \mathbf{\xi}_{N} \\
 \Delta \mathbf{d}^{-1}_i
\end{array}\right]
\end{aligned}
\end{equation}

We use the Gauss-Newton method to simultaneously optimize both the pose and the inverse depth. $\Delta \mathbf{\xi}_i$ represents a update of camera pose and $\mathbf{C}_{ii}$ is a diagonal matrix, the rest variables are sub-blocks of the matrix in Eq.~\ref{hessian matrix}.

To eliminate the depth state, we calculate the Schur Complement of the Hessian with respect to $\mathbf{C}$, which effectively constructs an inter-frame pose constraint containing the linearized BA information. The calculations in Eq.~\ref{schur complement} can be efficiently parallelized on the GPU:

\begin{equation}
\label{schur complement}
(\mathbf{B}_i - \mathbf{E}_i \mathbf{C}_i^{-1}\mathbf{E}_i^T)\Delta  \mathbf{\xi}_{i,1,...,N} = \mathbf{v}_i - \mathbf{E}_i \mathbf{C}_i^{-1} \mathbf{z}_i.
\end{equation}
where $\mathbf{B}_i, \mathbf{E}_i, \mathbf{C}_i, \mathbf{v}_i, \mathbf{z}_i$ are blocks in Eq.~\ref{hessian matrix}. After updating the pose, we can then update the inverse depth state using:
\begin{equation}
\Delta (\mathbf{d}_i^{-1}) = \mathbf{C}_i^{-1} \left( \mathbf{z}_i - \mathbf{E}_i^T  \Delta  \mathbf{\xi}_{i,1,...,N}  \right).
\end{equation}

We use the convex upsampling method from DROID-SLAM, as defined in RAFT, to obtain full-resolution depth. This upsampling approach takes a convex combination of the neighboring depth values, with the upsampling weights estimated by the GRU network.


\subsection{Visual Inertial Factor Graph}\label{subsec:vi_graph}

The factor graph we constructed includes visual factors and optionally IMU pre-integration factors, the factor graph optimization is implemented via GTSAM~\cite{gtsam}. The state variables in the IMU pre-integration factor are:

\begin{equation}
\mathbf{b}_{k} = \begin{bmatrix} \mathbf{b}_{a,k} & \mathbf{b}_{g,k} \end{bmatrix}, 
\mathbf{x}_{k} = \begin{bmatrix} \mathbf{T}_{b_{k}}^w & \mathbf{v}_{b_{k}}^w & \mathbf{b}_{k} \end{bmatrix}.
\end{equation}

Where $\mathbf{T}_{b_{k}}^w$ represents the IMU pose in the world frame, $\mathbf{v}_{b_{k}}^w$ represents the velocity, and $\mathbf{b}_{a,k}, \mathbf{b}_{g,k}$ are the accelerometer and gyroscope biases.
The visual factor is a pose constraint derived through Schur elimination as described above. Note that the poses in our factor graph are defined as transformations from the IMU frame to the world frame. Our visual factor is expressed as:

\begin{equation}
\begin{aligned}
\mathbf{T}_w^{c_k} &=  (\mathbf{T}_{b_k}^w \mathbf{T}_{c}^b)^{-1} \\
\mathbf{X}_c &= \begin{bmatrix} {\mathbf{\xi}_{w}^{c_0}}^T & {\mathbf{\xi}_{w}^{c_1}}^T & ... & {\mathbf{\xi}_{w}^{c_k}}^T \end{bmatrix}^T  \\
\mathbf{E}_c(\mathbf{X}_c) &= \frac{1}{2} \mathbf{X}_c^T \mathbf{H}_c \mathbf{X}_c - \mathbf{X}_c^T \mathbf{v}_c .
\end{aligned}
\end{equation}

where  $\mathbf{T}_{c}^b$ denotes the camera-to-IMU extrinsic transformation, $\mathbf{\xi}_w^{c_k}$ is Lie algebra of $\mathbf{T}_{w}^{c_k}$, $\mathbf{H}_c,\mathbf{v}_c$ are the information matrix and vector during DBA.

We follow the method of~\cite{imu-preintegrate} to compute IMU pre-integration. The residual for preintegrated IMU measurement can be defined as $\mathbf{r}_{b}\left(\mathbf{x}_{k}, \mathbf{x}_{k+1}\right)$: 
\begin{equation}
\begin{aligned}
\label{Gaussian_significance}
\begin{bmatrix}
 \textbf{R}_{w}^{b_k}\left(\mathbf{p}_{b_{k+1}}^{w}-\mathbf{p}_{b_{k}}^{w}+\frac{1}{2} \mathbf{g}^{w} \Delta t_{k}^{2}-\mathbf{v}_{b_{k}}^{w} \Delta t_{k}\right) - \hat{\mathbf{\alpha}}_{b_k}^{b_{k+1}} \\
\mathbf{R}_{w}^{b_k}\left(\mathbf{v}_{b_{k+1}}^{w}+\mathbf{g}^{w} \Delta t_{k}-\mathbf{v}_{b_{k}}^{w}\right)- \hat{\mathbf{\beta}}_{b_k}^{b_{k+1}}  \\

\mathrm{Log}(  (\mathbf{R}_{b_k}^w)^{-1}\mathbf{R}_{b_{k+1}}^w(\hat{\mathbf{\gamma}}_{b_{k+1}}^{b_{k}})^{-1} )
 \\
\mathbf{b}_{a, k+1}-\mathbf{b}_{a, k} \\
\mathbf{b}_{g, k+1}-\mathbf{b}_{g, k}
\end{bmatrix}
\end{aligned}
\end{equation}

Where $\mathbf{p}_{b_{k}}^{w}$ and $\textbf{R}^{w}_{b_k}$ represents the translation vector and rotation matrix of $\textbf{T}^{w}_{b_k}$, $\hat{\mathbf{\alpha}}_{b_k}^{b_{k+1}}, \hat{\mathbf{\beta}}_{b_k}^{b_{k+1}}, \hat{\mathbf{\gamma}}_{b_k}^{b_{k+1}} $ are the IMU pre-integration terms~\cite{VINS-MONO}, $\mathbf{g}^w$ is the gravity, $\Delta t_k$ is the time interval.

\subsection{Depth Uncertainty Estimation}\label{subsec:depth_uncert}

Our probabilistic depth uncertainty is inherently derived from the information matrix of the underlying Dense Bundle Adjustment (DBA) process. The primary goal of depth uncertainty is to suppress noise and mitigate artifacts. Using the sparse form of the Hessian matrix, we can calculate the marginal covariances for per-pixel inverse depth values. The marginal covariance of the inverse depth is formulated as:

\begin{equation}
\begin{aligned}
\label{eq:depth_uncert}
\mathbf{\Sigma}_T &= (\mathbf{H}/\mathbf{C})^{-1} \\
\mathbf{\Sigma}_{d^{-1}} &= \mathbf{C}^{-1} + \mathbf{C}^{-1} \mathbf{E}^T \mathbf{\Sigma}_T \mathbf{E} \mathbf{C}^{-1} \\
&= \mathbf{C}^{-1} + \mathbf{C}^{-1}\mathbf{E}^T(\mathbf{L}\mathbf{L}^T)^{-1}\mathbf{E} \mathbf{C}^{-1} \\
&= \mathbf{C}^{-1} + (\mathbf{L}^{-1}\mathbf{E}\mathbf{C}^{-1})^T (\mathbf{L}^{-1}\mathbf{E}\mathbf{C}^{-1}).
\end{aligned}
\end{equation}

where $\mathbf{L}$ is the lower triangular Cholesky factor. Finally, with all the information provided by the visual-inertial front end – including poses $\{T_t\}$, depths $\{D_t\}$, their associated depth uncertainties $\{U_t\}$, and the input RGB images $\{I_t\}$, we can incrementally construct and maintain our 2D Gaussian Map.

\section{2D Gaussian Map}\label{2d gaussian map}

We will first give a comprehensive introduction to the online mapping process, followed by a detailed explanation of the score manager, the sample rasterizer, and the pose refinement mechanism.

\subsection{Online Mapping Process}\label{subsec:only_mapping_process}

For the initialization of the mapping module, the 2D Gaussian Map is initialized after the VIO front end has processed the first batch of keyframes. For each frame $\{I_t, D_t, U_t, T_{c_t}^w\}$, pixels with excessive depth or high uncertainty are masked out. Then, $k$ points ($k=50,000$) are randomly sampled and projected to obtain point clouds in the world coordinate system. The Gaussian properties are initialized following the method described in 2DGS~\cite{2DGS}.  

\begin{equation}
\begin{aligned}
\label{Gaussian_property}
\mathcal{G} = \{ g_i:
 (\mu_i,r_i,\alpha_i,c_i) \mid \forall g_i \in \mathcal{G} \}.
\end{aligned}
\end{equation}

We follow the rendering approach of 2DGS to render color $C$, depth $D$, normal $N$ and accumulation $A$, as shown in Eq.~\ref{Eq:render}, where $z_i$ represents the depth value of the Gaussian center in the camera coordinate system, and $n_i$ represents the normal vector of the gaussian ellipsoid, with its positive direction aligned with the ray. For simplicity in notation, the rendering process (Eq.~\ref{Eq:render}) will be represented as $\mathcal{R}(\cdot)$.

\begin{equation}
\begin{aligned}
\label{Eq:render}
f(p) &= \mathrm{\alpha} \cdot \mathrm{exp}(-\frac{1}{2}(p-\mu)^T(p-\mu)) \\
(C,D,N,A) &= \sum_{i=1}^N (c_i,z_i,n_i, 1)f_i  \Pi_{j=1}^{i-1} (1-f_j) .
\end{aligned}
\end{equation}

The Mapping module and the VIO Front End operate as two parallel threads. For the subsequent incremental mapping process, we do not adopt the original 3DGS clone-and-split strategy for densification because we found in practice that the reset opacity operation performs unsatisfactorily in the GS-SLAM setting. Instead, we demonstrate that adding a relatively large number of Gaussians first and then pruning unnecessary ones is highly effective.

When the front end adds a new keyframe $\textbf{T}_{c_t}^w$, new Gaussian ellipsoids are added before training. First, the color and depth of $\textbf{T}_{c_t}^w$ are rendered. Then, two operations are performed: deleting conflicting Gaussians and adding necessary ones. Gaussians with high depth or RGB losses within the view frustum are removed, as well as those with excessively large projection radii. This is determined by projecting each Gaussian’s center onto the image. After deletion, the accumulation map is re-rendered, and new Gaussians are added in regions with low accumulation based on depth information. The number of new Gaussians is proportional to the area of low-accumulation pixels relative to the total pixel area. This redundant addition followed by selective pruning ensures robust performance, outperforming the original densification method in GS-SLAM, especially in forward-view scenarios like driving.

After adding the new Gaussian ellipsoids, we randomly sample frames from the latest keyframe list in the VIO Frontend for training. During training, the loss function $\mathcal{L}$ is computed according to 2DGS~\cite{2DGS}, as shown in Eq.~\ref{Eq:loss}. We add an additional accumulation loss to ensure that the masked-out regions do not contain black Gaussians.  During each training iteration, we record and update the local contribution score and error score of each Gaussian within the current keyframe list. These variables are used to maintain the Gaussians, which will be explained in detail in the next subsections.

\begin{equation}
\begin{aligned}
\label{Eq:loss}
\mathcal{L} &= \mathcal{\lambda}_{rgb}\mathcal{L}_{rgb} + \lambda_{depth}\mathcal{L}_{d} + \lambda_{norm}\mathcal{L}_{n}+\lambda_{acc}\mathcal{L}_{acc}.
\end{aligned}
\end{equation}

\subsection{Score Manager}\label{subsec:score_manager}

We propose a scoring mechanism to manage each Gaussian ellipsoid. This management involves status control (stable/unstable), storage control, and GPU-CPU transfer. The detailed algorithm flow is illustrated in Algorithm.~\ref{alg:gaussian_management}.

For a given keyframe list, we define a contribution score and an error score for each Gaussian. Our goal is to ensure that each Gaussian achieves the highest contribution score while causing the smallest error score. In a set of keyframes, a Gaussian contributes a weight to every pixel it touches. As shown in Eq.~\ref{Eq:Gaussian_significance}, for a Gaussian $g$, we accumulate the weights over $P$ pixels it touches to compute its total contribution to frame $t$. We then sum these contributions over the $K$ keyframes in the list to obtain the Gaussian's total contribution score $S_C(g)$. Each pixel has a loss value $L_{rgb}(u)$, and we compute the weighted sum of pixel losses over the PP pixels the Gaussian touches to get its total loss for frame $t$. Finally, we select the highest error score among the K keyframes in the list as the Gaussian's error score $S_E(g)$.

\begin{equation}
\begin{aligned}
\label{Eq:Gaussian_significance}
S_C(g) &= \sum_{t=0}^{K} \sum_{u=0}^{P}  f_i  \Pi_{j=1}^{i-1} (1-f_j) \\
S_E(g) &= \mathrm{max}(\{  \sum_{u=0}^{P}  \mathcal{L}_{rgb}(u) f_i  \Pi_{j=1}^{i-1} (1-f_j) \}_{t=0,...,K}) \\
ID(g) &= \mathrm{argmax}_t(\{ \sum_{u=0}^{P}  f_i  \Pi_{j=1}^{i-1} (1-f_j) \}_{t=0,...,K}). 
\end{aligned}
\end{equation}

The calculation logic for these two scores is consistent. Gaussians with high scores generally fall into three scenarios: some contribute significantly to a small number of frames while touching only a few frames, others have a relatively small impact on individual frames but influence a large number of frames, and some have a large impact on specific frames while also affecting many frames. When preserving Gaussians with high $S_C$, it is important to retain Gaussians from all these scenarios. However, when removing Gaussians with high $S_E$, we should avoid cases where a Gaussian causes minimal loss in each frame but accumulates a large $S_E$ due to being observed in many frames. To address this, we compute contribution scores using a summation, while error scores are computed using the maximum value. Furthermore, while calculating the contribution score, we also compute each Gaussian's contribution to individual frames. To capture this, we introduce a new variable for each Gaussian, denoted as ${ID}(g)$. This variable represents the specific frameID where the contribution of Gaussian $g$ is the highest. After calculating $S_E$, $S_C$, and ${ID}$, we manage the Gaussians based on these variables and the current pose through three processes: status control, storage control, and GPU-CPU transfer.

During status control, we handle the transitions of Gaussians on the GPU between two states: stable and unstable. The purpose of defining these two states is to speed up optimization by masking unstable Gaussians during sparse Adam updates and to identify which Gaussians should be removed during storage control. Every $\Delta n_{\text{status}}$ ($=400$) iterations, which corresponds to a full replacement of the local keyframe set, unstable Gaussians with contribution scores $S_C$ lower than ${S}_C^{status} (=10^{-4})$ are transitioned to the stable state, ensuring that unnecessary Gaussians are effectively removed, thereby reducing storage and computational overhead. Conversely, stable Gaussians with error scores $S_E$ exceeding a predefined threshold ${S}_E^{status} (=0.5)$ are transitioned to the unstable state, with their $S_C$ and $S_E$ reset to zero. This ensures that if the system revisits previously explored areas, these stable Gaussians can be reintroduced into the optimization process, allowing for further refinement of historical Gaussians when they become relevant again.

During storage control, we remove unnecessary Gaussians to optimize memory and computational resources. Every $\Delta n_{\text{storage}}$ ($=200$) iterations, we prune Gaussians with contribution scores $S_C$ below a certain threshold $S_C^{storage} (=0.5)$  and a status marked as unstable. This approach is necessary because, in experiments involving multi-room environments or complex road structures, we noticed that previously visited positions (historical Gaussians) can reappear in the view frustum. However, due to occlusion or distance, these Gaussians have a low contribution ($S_C$) to the current keyframe list. When pruning, it’s important to distinguish these historical Gaussians within the frustum from those with genuinely low contribution. Relying solely on projection radius, distance, or $S_C$ is insufficient and risks pruning important historical Gaussians, which would be disastrous. Our stable status resolves this issue by effectively preserving these Gaussians. This is particularly relevant in turning points, where historical Gaussians often re-enter the view frustum. In addition, storage control is highly effective and adaptable, and it can be applied to all Gaussian-based methods. According to our experiments, using storage control to reduce the number of Gaussian ellipsoids can cut their count by half without compromising rendering quality.

During GPU-CPU transfer, we address the memory limitations of GPU when handling large-scale street-level scenes containing tens of millions of Gaussians. We transfer Gaussians between CPU memory (RAM) and GPU memory. Every $\Delta K$ ($=8$) keyframes, Gaussians are transferred between CPU and GPU memory based on their distance from the current pose. To reduce computational overhead, we use the pose index $\mathrm{ID}(g)$ to calculate distances between poses instead of directly computing distances for all Gaussian centers. Gaussians linked to poses within a specified distance threshold $\tau$ are transferred from CPU to GPU memory for faster access, while those beyond $\tau$ are moved from GPU to CPU memory and removed from GPU storage. This strategy dynamically balances memory usage, ensuring relevant Gaussians stay on the GPU while less relevant ones are offloaded to the CPU, maintaining both efficiency and rendering quality.

\begin{algorithm}
\caption{Score Manager}
\label{alg:gaussian_management}
\begin{algorithmic}[1]
\STATE \textbf{Input:} $\Delta n_{\text{status}} = 400$, $\Delta n_{\text{storage}} = 200$, $\Delta K = 8$
\STATE \textbf{Thresholds:} $S_C^{\text{status}} = 10^{-4}$, $S_E^{\text{status}} = 0.5$, $S_C^{\text{storage}} = 0.5$, distance threshold $\tau$
\STATE \textbf{Initialize:} $j = 0$ 
\FOR{each keyframe $k$}
    \FOR{each iteration $i$}
        \STATE $j = j + 1$
        \IF{$j \mod \Delta n_{\text{storage}} = 0$}
            \FOR{each Gaussian $g$}
                \IF{$g$ is unstable \AND $S_C(g) < S_C^{\text{status}}$}
                    \STATE Set $g$ to stable
                \ELSIF{$g$ is stable \AND $S_E(g) > S_E^{\text{status}}$}
                    \STATE Set $g$ to unstable, reset $S_E(g)$, $S_C(g)$
                \ENDIF
            \ENDFOR
        \ENDIF
        \IF{$j \mod \Delta n_{\text{storage}} = 0$}
            \FOR{each Gaussian $g$}
                \IF{$g$ is unstable \AND $S_C(g) < S_C^{\text{storage}}$}
                    \STATE Prune $g$
                \ENDIF
            \ENDFOR
        \ENDIF
        \IF{$k \mod \Delta K = 0$}
            \FOR{each Gaussian $g$}
                \IF{Pose distance $d(T_{ID(g)}) < \tau$}
                    \STATE Transfer $g$ to GPU storage
                \ELSE
                    \STATE Move $g$ to CPU, remove from GPU storage
                \ENDIF
            \ENDFOR
        \ENDIF
    \ENDFOR
\ENDFOR
\end{algorithmic}
\end{algorithm}

\subsection{Sample Rasterizer}\label{subsec:sample_rasterizer}

\begin{figure}[htbp]
    \centering
    \subfloat[2DGS Backward Process.]{%
        \includegraphics[width=1.0\linewidth]{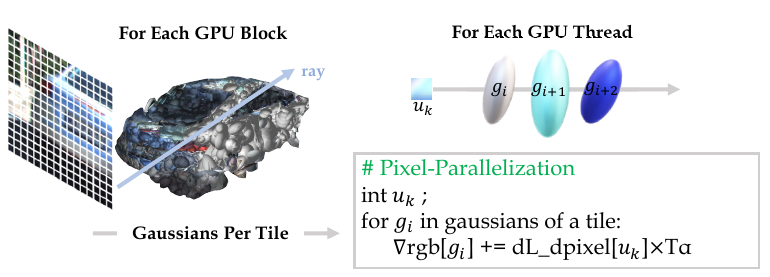} 
        \label{SampleRasterizer_subfigA}
    }%
    \vspace{1em} 
    \subfloat[Ours Backward Process.]{%
        \includegraphics[width=1.0\linewidth]{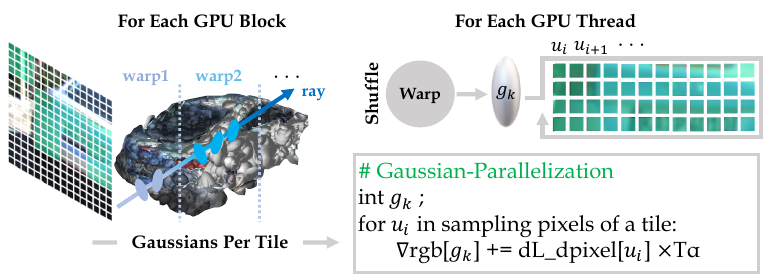} 
        \label{SampleRasterizer_subfigB}
    }%
    \caption{\textbf{Sample Rasterizer.} In our backpropagation process, each thread is responsible for one Gaussian, and the number of iterations depends on the number of sampled pixels.}
    \label{SampleRasterizer_Pipeline}
\end{figure}

In the original Gaussian Splatting method, the process for backpropagation mirrors the forward propagation in a symmetrical fashion. Each GPU block is responsible for one tile, with each tile containing 16$\times$16 pixels. The 256 threads within a block each handle the backpropagation for one pixel, specifically propagating the loss $L_i$ of pixel $p_i$ back to the Gaussians corresponding to that pixel. Consequently, the number of iterations each thread performs depends on the number of Gaussians associated with that pixel. This results in a bottleneck effect, where the overall backpropagation time is determined by the pixel with the maximum number of Gaussians.

Inspired by Taming 3DGS~\cite{taming3dgs} and prior NeRF approach~\cite{NeRF}, which achieve backpropagation by sampling pixels, we examined why pixel sampling did not accelerate backpropagation in the vanilla Gaussian Splatting's rasterization pipeline. The issue was that pixel sampling merely reduced the number of active threads per block, without changing the number of iterations each thread performs (still dependent on the number of Gaussians associated with each pixel).

To address this, we introduced a modification during forward propagation. As shown in Fig.~\ref{SampleRasterizer_Pipeline}, for each thread, we store intermediate variables in a buffer at intervals of 32 Gaussians. During backpropagation, we divided the GPU into multiple warps for computation, where each warp, consisting of 32 threads, performs backpropagation on the Gaussians within the warp. This change reduces the number of iterations per thread to match the number of pixels associated with the current tile. We further optimized by selecting a subset of pixels with the highest loss rates $r$ within each tile for backpropagation. With this approach, each thread’s iteration count is reduced to $256 \times r$. In our experiments, we set $r = 0.5$, which resulted in a backpropagation speedup of $273\%$ compared to the original method. Detailed experimental results are documented in Sec.~\ref{subsubsec:sample_rasterizer}.

\subsection{Single-to-Multi Pose Refinement}\label{subsec:pose_refine}
Existing GS-based SLAM~\cite{MONO-GS,SplaTAM,GS-SLAM}, optimizes localization by propagating gradients to the positional property of Gaussians, which then pass these gradients on to the current frame's pose. However, this method is relatively inefficient. In Eq.~\ref{Eq:Gaussian_significance}, we associated Gaussians with their respective keyframes. Based on this, we implemented a system where gradients of different Gaussian poses are propagated as pairs to different keyframe poses, thereby enabling the rendering of a single frame to optimize multiple frame poses. The optimized poses replace the visual frontend's pose buffer, facilitating further rounds of optimization.

As in Eq.~\ref{Eq:pose_refine}, for the $k$th keyframe, the pose is represented as $\textbf{T}_{c_k}^{w}$. From $S_C$, we obtain the subset of Gaussians associated with this frame, denoted as $\{g_{c_k}\}$. We introduce the camera pose transformation matrix $\textbf{T}_{c_k}^{\hat{c_k}}$ as an optimization variable. Subsequently, we render RGB image $\hat{I_k}$ and perform backpropagation to optimize the poses of all keyframes within the visible range by minimizing the rendering loss. This process effectively adjusts the keyframe poses based on their rendering performance to enhance the overall quality of the visualization.

\begin{equation}
\begin{aligned}
\label{Eq:pose_refine}
\hat{\mu_k} &= \textbf{T}_{c_k}^{w} \textbf{T}_{c_k}^{\hat{c_k}}  \textbf{T}_{w}^{c_k} \mu_k,  \hat{\textbf{T}_{c_k}^w} =\textbf{T}_{c_k}^w(\textbf{T}_{c_k}^{\hat{c_k}})^{-1}\\
\hat{I_k} &= \mathcal{R}(\{\hat{\mu_k}, s_k, c_k, r_k\}, \hat{\textbf{T}_{c_k}^w}) \\
& \mathrm{min}_{\{\textbf{T}_{c_k}^{\hat{c_k}}\}} \mathcal{L}_{rgb} (\hat{I_k}, I_k) .
\end{aligned}
\end{equation}

\section{NVS Loop Closure}\label{nvs loop closure}
In monocular setups that lack scale information,  loop closure is essential to eliminate accumulated errors, especially in large-scale environments. We propose a novel Gaussian Splatting based loop detection and correction method. Instead of using the Bag of Words (BoW) approach for loop detection, we leverage the novel view synthesis capabilities of gaussian splatting from new viewpoints to determine if a loop has been detected (Sec.~\ref{SeC:loopdetection}). Following this, we use graph optimization to correct poses and use gaussians' frame index $ID(g)$ to correct the 2D Gaussian Map (Sec.~\ref{loopcorrection}).

\begin{figure*}[!t]
\centering
\includegraphics[width=0.95\textwidth]{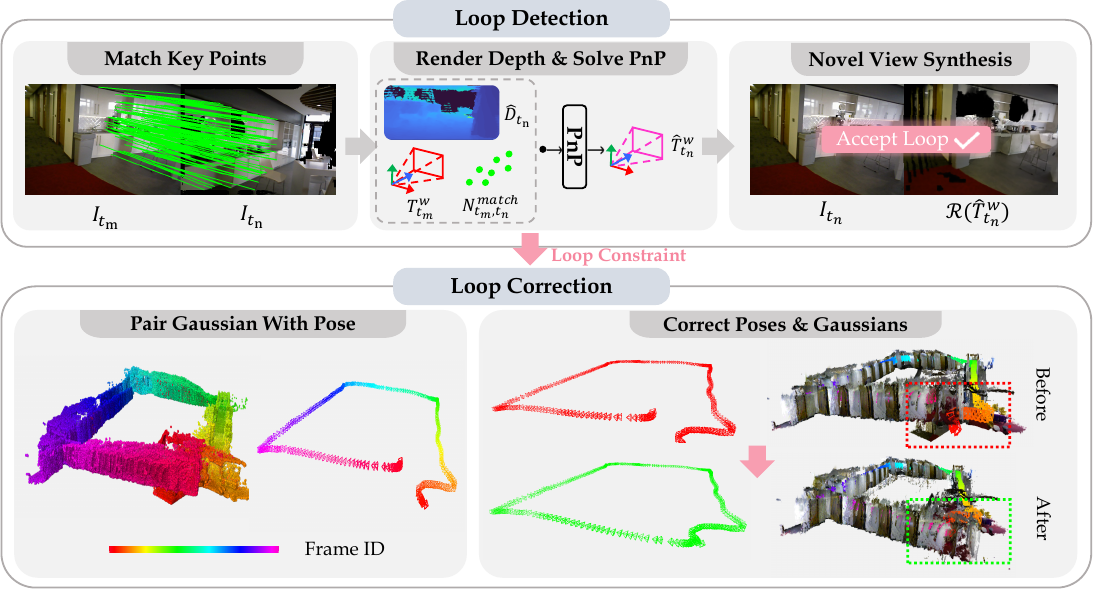}
\caption{\textbf{Pipeline of NVS Loop Closure.} We perform feature matching, filtering, and novel view synthesis on keyframes that meet the distance threshold requirements to achieve loop detection. Once a loop is detected, we implement loop correction of the pose and Gaussian map through pairwise Gaussian with pose alignment and graph optimization.}
\label{Fig:pipeline_of_nvs_loop_closure}
\end{figure*}

\subsection{Loop Detection}\label{SeC:loopdetection}

As illustrated in the upper section of Fig.~\ref{Fig:pipeline_of_nvs_loop_closure}, our loop detection process comprises three key steps: matching feature points with historical frames, deriving the relative poses of the two frames from the matched feature points and rendered depths, and synthesizing a novel view using the new poses to ascertain the presence of a loop closure.

\subsubsection{Match Key Points}
We extract and match feature points ~\cite{lightglue} with historical frames $\{I_{t_k}\}$ located within a specified range of the current pose $T_{{t_n}}^{w}$ and with a frame ID difference exceeding ten from the current frame. The number of feature points successfully matched between historical frame $I_{t_k}$ and current frame, denoted by $N_{match}(t_k, t_n)$, is systematically recorded. Frames whose match counts exceed the threshold $N_{match}^{th} (=50)$ are then organized in descending order based on the number of matches and we denote this set as $\{I_{t_k}\}^{filt}$.

\subsubsection{Render Depth \& Solve PnP}
We sequentially check $\{I_{t_k}\}^{filt}$ in descending order based on the number of keypoints. First, we use $T_{t_n}^w$ to render the depth map $\hat{D_{t_n}}$ of current frame. For frame $I_{t_m} \in \{{I_{t_k}}^{filt}\}$, we perform a Perspective-n-Point (PnP) computation using the matched feature points $N_{t_m,t_n}^{match}$ to estimate the relative pose $\hat{T_{t_n}^{t_m}}$ between $t_m$ and $t_n$. Subsequently, the global pose is computed as $\hat{T^{w}_{t_n}} = T_{t_m}^w \hat{T_{t_n}^{t_m}} $. It is important to note that, due to the instability of PnP when using feature points with distant depth values, we restrict the selection to points with depth values below a fixed threshold.

\subsubsection{Novel View Synthesis}
The core of the loop detection problem is determining whether two images capture the same scene. With the inherent novel view synthesis capability of 3DGS, this problem transforms into verifying whether the newly captured image can serve as a novel viewpoint of the Gaussian Map. Loop detection can be directly assessed by calculating the L1 Loss between the newly synthesized view $\mathcal{R}(\hat{T^{w}_{t_n}})$ and the original image $I_{t_n}$. The criterion is as follows: if the color loss is below a specific threshold or less than one-tenth of the median color loss among other frames in $\{I_{t_k}\}^{\text{filt}}$, a loop is considered detected.

\subsection{Loop Correction}\label{loopcorrection}
After detecting a loop closure and the corresponding loop closure constraints, we correct the poses of historical keyframes and the Gaussian Map. It is not feasible to directly optimize the Gaussian Map based on loop closure constraints because, in large-scale environments, loop closure errors tend to be significant. Directly retraining the Gaussian Map with corrected poses may not converge effectively. Therefore, we first associate each Gaussian with a historical keyframe pose to ensure consistency between our Gaussian Map and poses. On this basis, we then proceed with fine-tuning.

\subsubsection{Pair Gaussian with Pose}
For all historical keyframes, we forward propagate and record the contribution score of each Gaussian in each frame sequentially. Each Gaussian selects the pose corresponding to its highest score as the matched pose. Due to the extremely fast rendering speed, this process takes approximately two seconds for one thousand frames.

\subsubsection{Correct Pose \& Gaussians}

We construct the pose graph for all historical frames $\{{T_{c_k}^w}\}$ and add loop closure constraints to it. Then, we perform graph optimization to obtain the updated global poses of the historical keyframes $\{{T_{c_k}^w}'\}$. For each historical keyframe, we calculate its scale using the ratio of the translation vector norms before and after the transformation. As described in Eq.~\ref{Eq:loop_correction}, $k$ is $ID(g_i)$ and $R(\cdot)$ represents the transformation from a quaternion to a rotation matrix. We compute each Gaussian's new position ${\mu}'$ and rotation $r_i$, while keeping other attributes unchanged. Subsequently, we retrain the model for one hundred iterations on the global set of historical keyframes and record the $S_C$. Finally, we perform an additional step to prune Gaussians based on $S_C$ to further optimize storage overhead.

\begin{equation}
\begin{aligned}
\label{Eq:loop_correction}
{\mu_i}' &= {T_{c_k}^w}' T_{w}^{c_k} \mu_{i} \\
{r_i}' &=  R^{-1}({T_{c_k}^w}' T_{w}^{c_k} R(r_{i})).
\end{aligned}
\end{equation}

\section{Dynamic Object Eraser}\label{dynamic object eraser}

The underlying assumption of Gaussian Splatting is that scenes are static. However, in real-world applications, especially in large-scale environments, dynamic distractors like vehicles or pedestrians are common. Previous dynamic Gaussian Splatting  methods~\cite{4dgs, dynamic-3dgs, realtime4dgs, pvg} were implemented in offline training settings. These approaches model the 4D space and train the relationships between Gaussian properties and time across the entire dataset in an offline manner. However, such methods are not suitable for SLAM, which requires incrementally loading data. Considering that SLAM’s mapping is an online process and that Gaussian Splatting has the capability for novel view synthesis, we designed a heuristics-guided segmentation method to distinguish masks of dynamic objects.

First, we apply an accelerated open-set semantic segmentation model~\cite{fastsam} on the entire image to generate a set of semantic masks, denoted as $\{M_{k}\}_{k=0,1,...,K}$. When a new keyframe $I_{t}$ arrives, we render the color of current frame $R(T_{c_t}^w)$ before adding new Gaussians. Next, we calculate the SSIM loss and the L1 loss separately with respect to the new keyframe. We observed that SSIM is particularly sensitive to textures, whereas the L1 loss is more sensitive to color value differences. By multiplying these two losses, we compute the re-rendering Loss, $\mathcal{L}_{\text{re}}$.  Note that this loss is initially calculated at pixel level before taking the overall average, we denote the 90\% percentile of this pixel-level loss as $\mathcal{L}_{\text{re}}^{90\%}$. However, for dynamic objects with relatively smooth textures, the re-rendering loss is only primarily noticeable around the edges. To address this issue, we modify the re-rendering loss calculation by incorporating depth uncertainty as mention in Eq. \ref{eq:depth_uncert}, $\mathcal{L}_{\text{re}} = \mathcal{L}_{\text{SSIM}} \cdot \mathcal{L}_1 \cdot \Sigma_{d^{-1}}$.  This enables us to more effectively identify and determine the mask for moving objects $M_{\text{dyn}}$.

\begin{equation}
\begin{aligned}
\label{Gaussian_significance}
M_{dyn,k} &= (\frac{\sum \textbf{1}(\mathcal{L}_{re}(M_{k})>\mathcal{L}^{90\%}_{re})}  {\sum \textbf{1}(M_{k})}>\gamma)\wedge (\overline{{\mathcal{L}_{re}}}(M_{k}) >\mathcal{L}_{re}^{th})\\
M_{dyn} &= \bigcup_{k=0}^{K} M_{dyn,k}.
\end{aligned}
\end{equation}

\begin{figure}[!t]
\centering
\includegraphics[width=0.5\textwidth]{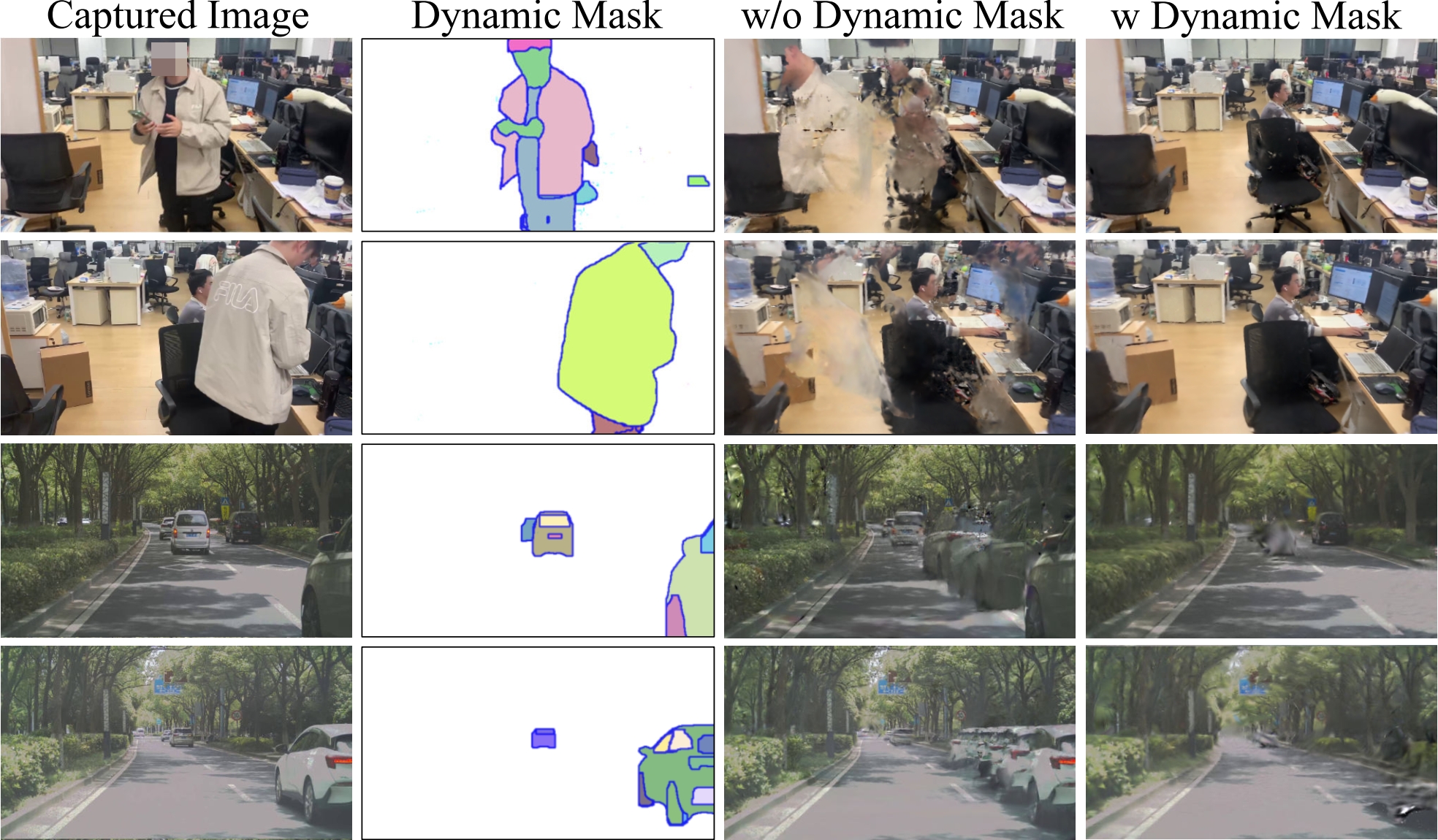}
\caption{\textbf{Effect of Dynamic Object Eraser.} Our dynamic eraser can filter out moving people indoors and fast-moving vehicles outdoors, preventing the Gaussian map from being affected by dynamic floaters.}
\label{pipeline}
\end{figure}

Where $\sum \textbf{1}(\cdot)$ represents pixel number of the mask and $\gamma$ is set to $20\%$.
We identify a semantic mask as a dynamic mask when the rerendering loss in the region covered by the mask exceeds a certain threshold in more than $\gamma$ of the mask's pixels, and the color rendering loss is relatively high. We filter them out during the addition of new Gaussians and in the subsequent rendering process.

\section{Experimental Evaluation}

We conducted comprehensive comparative experiments to evaluate  our framework's tracking performance and mapping performance emphasized by Gaussian Splatting. Additionally, we carried out comparative experiments on the dynamic object eraser and performed ablation studies on the individual components of our system. Finally, we analyzed the runtime of the system and introduced the mobile app we developed, along with real-world experiments.

\subsection{Experimental Setup}

\subsubsection{Datasets and Metrics}
To validate the effectiveness and robustness of our algorithm, we exclusively use real-world datasets rather than simulated data. Our experiments include two large-scale indoor scenarios, a classic dynamic indoor SLAM dataset, and five different outdoor scenarios characterized by varying lighting conditions, movement speeds, and capture devices. Additionally, we collect real-world data using consumer-level smartphone’s sensors.

For indoor scenarios, we evaluated ScanNetV1~\cite{scannet} and BundleFusion~\cite{bundlefusion}. 
ScanNetV1~\cite{scannet} is a widely used RGB-D dataset in SLAM research, offering over 1500 indoor scenes with 3D camera pose annotations. We selected six large-scale scenes with significant lighting variations for our experiments. The official BundleFusion dataset was tested on five scenes (apt0, apt2, copyroom, office0, office2), with reference trajectories provided by the official dataset. 

For outdoor scenarios, we conducted experiments on three driving-view datasets: KITTI~\cite{KITTI}, KITTI-360~\cite{KITTI360}, and Waymo~\cite{Waymo2021Tech}, a drone-view dataset: MegaNeRF~\cite{meganerf}, and a cycling-view dataset: Hierarchical dataset~\cite{hierarchical_3dgs}. The KITTI dataset was collected in urban, rural, and highway settings operating at 10 Hz, along with LiDAR scans captured by a Velodyne HDL-64 and ground-truth trajectories recorded by an OXTS 3003 GPS/IMU. The KITTI-360~\cite{KITTI360} dataset consists of 9 sequences covering over 73 km, with data captured by a stereo pair at 10 Hz (rectified resolution: 1408×376) and ground-truth poses derived through large-scale optimization combining OXTS data, laser scans, and multi-view images. The Waymo~\cite{Waymo2021Tech} dataset was collected in San Francisco, Mountain View, and Phoenix using five LiDAR sensors and five high-resolution cameras (rectified resolution: 1920×1280) operating at 10 Hz, with official vehicle poses provided for each range image. For MegaNeRF~\cite{meganerf}, we evaluated two scenes: Building, which features grid-pattern footage of a 500×250$m^2$ industrial area, and Rubble, both with GPS-derived camera poses. The Hierarchical 3DGS~\cite{hierarchical_3dgs} dataset includes walking and cycling data; the Campus subset was captured with five GoPro HERO 6 cameras (resolution: 1444$\times$1080), while the Small City subset was recorded at 7 km/h. Reference trajectories were generated using COLMAP~\cite{colmap} with a hierarchical mapper and per-chunk bundle adjustment.

To assess global consistency between estimated and ground-truth trajectories, we calculate Absolute Trajectory Error (ATE)~\cite{atebenchmark} for indoor datasets and Relative Pose Error (RPE) following~\cite{KITTI} for large scale outdoor datasets via the evo toolkit~\cite{grupp2017evo}. For rendering quality, we applied metrics like Peak Signal-to-Noise Ratio (PSNR), Learned Perceptual Image Patch Similarity (LPIPS), and Structural Similarity Index (SSIM).

\subsubsection{Parameters and Implementation Details}

All experimental results were recorded using a single RTX 4090 GPU and an Intel Xeon 6133 CPU (2.50GHz). The preset parameters are divided into three components: the front end, Gaussian maps, and loop closure detection.

For the VIO front end, we set the optical flow threshold between two frames to be greater than 2.4 pixels for a new keyframe to be considered. The length of the local keyframe list is set to eight. For Gaussian maps, each keyframe in the ScanNet and BundleFusion datasets undergoes 80 iterations of rendering training and 10 iterations of pose optimization. For MegaNeRF, each new keyframe undergoes 50 iterations of rendering training without pose refinement. For other datasets and real-world experiments, 100 iterations of training are performed per keyframe. Additionally, in the monocular settings of real-world experiments, we use~\cite{yin2023metric3d} to obtain depth priors, enhancing the geometric quality of the results.

Following the advice of the original 3D Gaussian approach, we reduce the learning rate for positional attributes in outdoor scenes. The weights of the loss function are $\lambda_{rgb}=1.0, \lambda_{depth}=0.5, \lambda_{normal}=0.1, \lambda_{\alpha}=0.1$
For loop closure detection, the filtering radius is set to 15 meters for indoor scenes and 50 meters for outdoor scenes. The threshold for filtering by the number of matching points is set to 50, and the re-rendering loss threshold is set to 0.15.

Some methods in the comparison experiments face difficulties running successfully in outdoor scenarios. To ensure that the comparison experiments are reasonable and meaningful, we provide pseudo ground-truth depth or increase the number of iterations where necessary. Specific configurations will be explained in detail in the experimental section.

\subsection{Localization Performance}
In this section, we compare the pose estimation performance of VINGS-Mono with both traditional SLAM methods and Gaussian Splatting based methods.

\subsubsection{VO Comparison}
We conducted experiments on two large-scale indoor scene datasets including ScanNet~\cite{scannet} and BundleFusion~\cite{bundlefusion}, as well as two outdoor scene datasets Hierarchical~\cite{hierarchical_3dgs} and Waymo~\cite{Waymo2021Tech}. For our evaluation, we selected representative traditional SLAM methods ORB-SLAM3~\cite{ORBSLAM3_TRO} and DROID-SLAM~\cite{droid-slam}. To ensure a fair comparison for DROID-SLAM, we omitted the global BA post-processing step after the full run. It's worth noting that these methods do not have the capability to construct Gaussian maps. Additionally, we included one NeRF-based SLAM method~\cite{nerfslam} and two state-of-the-art monocular Gaussian-based SLAM methods MonoGS~\cite{MONO-GS} and Photo-SLAM~\cite{hhuang2024photoslam}. For indoor scenes, as shown in Tab.~\ref{Tab:indoor_vo}, our method performs on par with traditional SLAM methods but significantly outperforms existing monocular GS-based SLAM methods. 

\begin{figure*}[!t]
\centering
\includegraphics[width=\textwidth]{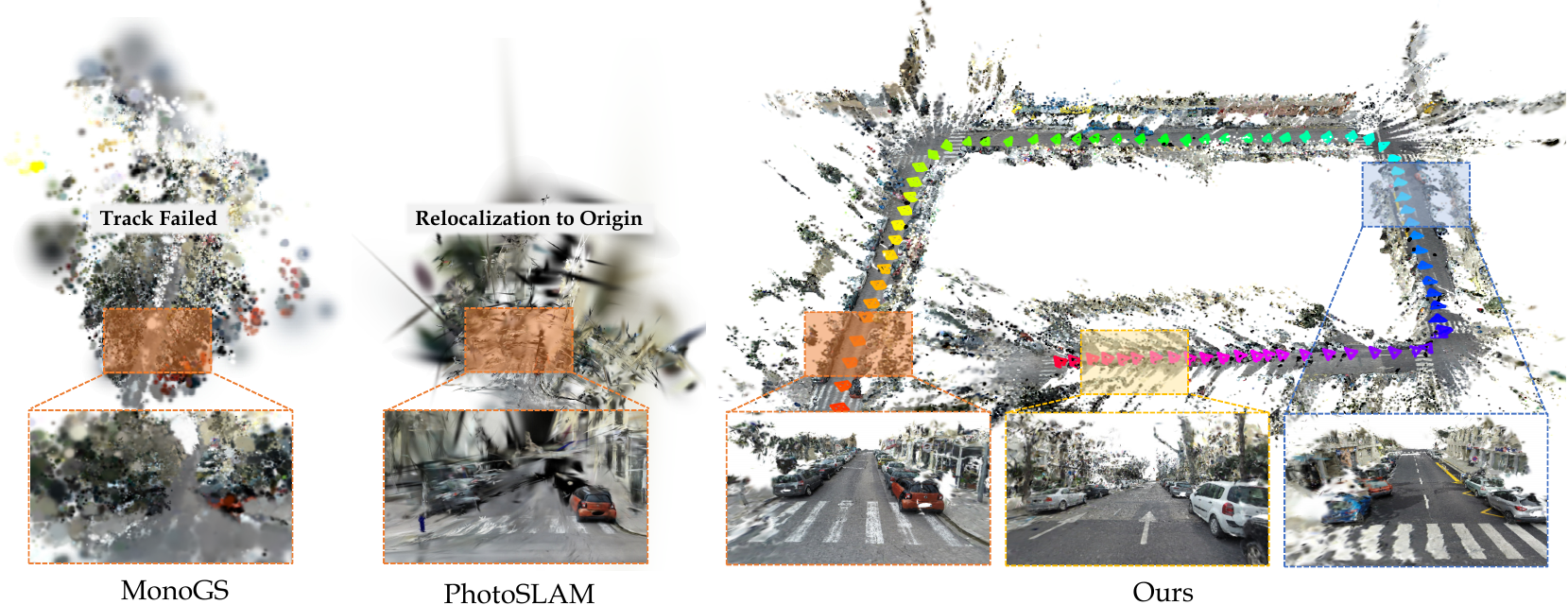}
\caption{\textbf{VO Performance on SmallCity of Hierarchical~\cite{hierarchical_3dgs}.}
MonoGS fails in tracking due to being obscured by large floaters, and Photoslam cannot match feature points to relocate to the starting point due to the lack of complex textures in and ego fast motion. In contrast, our method robustly and stably achieves localization and constructs high-quality Gaussian maps. }
\label{Fig:outdoor_vo_smallcity}
\end{figure*}

For the outdoor scenes which are our main focus, as shown in Tab.~\ref{Tab:outdoor_vo}, our method achieves better localization accuracy compared to existing approaches. Additionally, due to the faster movement speed (resulting in insufficient overlap between consecutive frames) and the large ground area in outdoor large-scale scenes like Hierarchical-Smallcity, which causes weaker textures, both ORB-SLAM and PhotoSLAM (which uses ORB as its frontend) relocated back to the origin. MonoGS, on the other hand, displayed a completely black image. All three methods successfully tracked less than half of the trajectory, as shown in Fig.~\ref{Fig:outdoor_vo_smallcity}. Therefore, for ORB-SLAM and PhotoSLAM, we only recorded the ATE for the first fifty frames and marked with an asterisk(*) in Tab~\ref{Tab:outdoor_vo}. However, our method was still able to handle large-scale scenes with relatively faster movement speeds effectively.

\begin{table*}
\centering
\caption{Monocular Localization results (ATE [cm]) on the indoor datasets ScanNet and BundleFusion. Red, orange, and yellow represent the \colorbox{red}{best}, \colorbox{orange}{second-best}, and \colorbox{yellow}{third-best} performance, respectively. For all evaluation scenarios, the same dataset with ground truth values was used as a reference to compute the average metrics.}
\label{Tab:indoor_vo}
\begin{tabular}{c|cccccc|ccccc} 
\hline
\multirow{2}{*}{ATE (cm) ↓} & \multicolumn{6}{c|}{ScanNet}                                                                                                                                                                                                  & \multicolumn{5}{c}{BundleFusion}                                                                                                                                                         \\
                          & 0054                                & 0059                               & 0106                               & 0169                               & 0233                               & 0465                                & apt0                               & apt2                                & copyroom                           & office0                            & office2                             \\ 
\hline
ORB-SLAM3                 & 243.26                              & 90.67                              & 178.13                             & 60.15                              & {\cellcolor[rgb]{1,0.6,0.6}}25.01  & 181.86                              & 89.38                              & {\cellcolor[rgb]{1,0.98,0.6}}148.04 & {\cellcolor[rgb]{1,0.6,0.6}}19.70  & {\cellcolor[rgb]{1,0.6,0.6}}31.41  & {\cellcolor[rgb]{1,0.98,0.6}}73.91  \\
DROID-SLAM                & 161.22                              & {\cellcolor[rgb]{1,0.98,0.6}}67.26 & {\cellcolor[rgb]{1,0.6,0.6}}11.20  & {\cellcolor[rgb]{1,0.98,0.6}}17.39 & 69.85                              & 116.42                              & {\cellcolor[rgb]{1,0.98,0.6}}87.37 & 265.64                              & {\cellcolor[rgb]{1,0.8,0.6}}27.59  & 116.33                             & {\cellcolor[rgb]{1,0.8,0.6}}49.32   \\
NeRF-SLAM                 & {\cellcolor[rgb]{1,0.98,0.6}}147.20 & {\cellcolor[rgb]{1,0.8,0.6}}26.95  & {\cellcolor[rgb]{1,0.98,0.6}}18.75 & {\cellcolor[rgb]{1,0.6,0.6}}13.53  & {\cellcolor[rgb]{1,0.8,0.6}}37.23  & {\cellcolor[rgb]{1,0.6,0.6}}73.32   & {\cellcolor[rgb]{1,0.8,0.6}}85.50  & 241.72                              & 59.20                              & {\cellcolor[rgb]{1,0.98,0.6}}59.08 & 83.57                               \\
MonoGS                    & {\cellcolor[rgb]{1,0.8,0.6}}70.189  & 97.24                              & 150.89                             & 191.98                             & 62.45                              & {\cellcolor[rgb]{1,0.98,0.6}}113.19 & 122.59                             & {\cellcolor[rgb]{1,0.8,0.6}}142.54  & 53.41                              & 62.67                              & 127.02                              \\
PhotoSLAM                 & 332.03                              & 205.01                             & 359.85                             & 151.61                             & 195.71                             & 294.20                              & 247.19                             & 320.91                              & 54.03                              & 271.87                             & 298.98                              \\
Ours                      & {\cellcolor[rgb]{1,0.6,0.6}}44.08   & {\cellcolor[rgb]{1,0.6,0.6}}15.96  & {\cellcolor[rgb]{1,0.8,0.6}}16.13  & {\cellcolor[rgb]{1,0.8,0.6}}16.84  & {\cellcolor[rgb]{1,0.98,0.6}}60.71 & {\cellcolor[rgb]{1,0.8,0.6}}92.83   & {\cellcolor[rgb]{1,0.6,0.6}}44.22  & {\cellcolor[rgb]{1,0.6,0.6}}136.69  & {\cellcolor[rgb]{1,0.98,0.6}}39.10 & {\cellcolor[rgb]{1,0.8,0.6}}44.44  & {\cellcolor[rgb]{1,0.6,0.6}}39.10   \\
\hline
\end{tabular}
\end{table*}

\begin{table}
\centering
\caption{Monocular Localization results (ATE [m]) on the outdoor datasets Waymo and Hierarchical3DGS. ``-'' indicates that the system failed to track in this scenario, ``*'' indicates only the first 50 frames were tested due to tracking failure.}
\label{Tab:outdoor_vo}
\begin{tabular}{c|ccc|cc} 
\hline
\multirow{2}{*}{RMSE [m] ↓} & \multicolumn{3}{c|}{Waymo}                                                                                & \multicolumn{2}{c}{Hierarchical3DGS}                                   \\
                         & Scene01                           & Scene03                           & Scene14                           & SmallCity                         & Campus                             \\ 
\hline
ORB-SLAM3                & {\cellcolor[rgb]{1,0.8,0.6}}1.21  & {\cellcolor[rgb]{1,0.6,0.6}}2.49  & {\cellcolor[rgb]{1,0.8,0.6}}2.48  & -                                 & -                                  \\
DROID-SLAM               & 2.38                              & {\cellcolor[rgb]{1,0.98,0.6}}2.94 & {\cellcolor[rgb]{1,0.98,0.6}}3.98 & 5.83                              & {\cellcolor[rgb]{1,0.98,0.6}}1.87  \\
 NeRF-SLAM                & {\cellcolor[rgb]{1,0.98,0.6}}2.05 & 5.87                              & 6.43                              & {\cellcolor[rgb]{1,0.8,0.6}}4.58  & {\cellcolor[rgb]{1,0.8,0.6}}1.44   \\
GO-SLAM                  & 3.15                              & 3.07                              & 5.13                              & {\cellcolor[rgb]{1,0.98,0.6}}5.79 & 3.50                               \\
MonoGS                   & 2.73                              & 10.73                             & 6.59                              & 6.05*                              & 20.81*                              \\
PhotoSLAM                & 3.15                              & 6.41                              & 7.30                              & 47.72*                             & 34.04*                              \\
Ours                     & {\cellcolor[rgb]{1,0.6,0.6}}0.91  & {\cellcolor[rgb]{1,0.8,0.6}}2.67  & {\cellcolor[rgb]{1,0.6,0.6}}2.27  & {\cellcolor[rgb]{1,0.6,0.6}}2.82  & {\cellcolor[rgb]{1,0.6,0.6}}1.03   \\
\hline
\end{tabular}
\end{table}

\subsubsection{VIO Comparison}

We compared the pure odometry accuracy on the competitive KITTI~\cite{KITTI} and KITTI360~\cite{KITTI360} datasets. Due to the significant storage and computational demands of kilometer-scale urban scenes, no existing NeRF/3DGS-based SLAM methods can run on both datasets. Therefore, we selected two feature-based methods, VINS-Mono and ORB-SLAM3, as well as two advanced learning-based methods, iSLAM~\cite{iSLAM} and Selective-VIO~\cite{selectiveVIO}, for comparison. To test whether our algorithm can robustly adapt to different IMU frequencies, we used the 10Hz KITTI sync data and the 100Hz KITTI360 unsync data. The original KITTI unsync data had issues with out-of-order timestamps. Following the recommendations in the issue reports, we organized and corrected the data. Notably, ORB-SLAM3 failed to run successfully on KITTI's Mono\&IMU configuration, so we recorded the results using the Stereo\&10Hz IMU setup instead. As shown in Tab. ~\ref{Tab:outdoor_vio}, for low-frequency IMU data, our method significantly outperforms feature-based SLAM algorithms and, in many scenarios, surpasses learning-based SLAM methods. Under the high-frequency IMU settings of the KITTI360 dataset, our method demonstrates a notably better localization performance compared to other approaches. Moreover, we are the first to propose a Gaussian Splatting-based VIO odometry.

\begin{table*}
\centering
\caption{Visual inertial localization results ($t_{rel}$ in $\%$ and $r_{rel}$ in $^{\circ} /100m$) on KITTI and KITTI360.}
\label{Tab:outdoor_vio}
\begin{tabular}{c|cccccccccc|cccccccccc} 
\hline
\multirow{3}{*}{$t_{rel}$↓ $r_{rel}$↓} & \multicolumn{10}{c|}{KITTI Sync}                                                                                                                                                                                                                                                                                                                                                                    & \multicolumn{10}{c}{KITTI360 Unsync}                                                                                                                                                                                                                                                                                                                                                                 \\
                                   & \multicolumn{2}{c}{02}                                                      & \multicolumn{2}{c}{06}                                                      & \multicolumn{2}{c}{07}                                                      & \multicolumn{2}{c}{08}                                                      & \multicolumn{2}{c|}{09}                                                     & \multicolumn{2}{c}{00}                                                      & \multicolumn{2}{c}{02}                                                      & \multicolumn{2}{c}{05}                                                      & \multicolumn{2}{c}{06}                                                      & \multicolumn{2}{c}{10}                                                       \\
                                   & $t_{rel}$                              & $r_{rel}$                              & $t_{rel}$                              & $r_{rel}$                              & $t_{rel}$                              & $r_{rel}$                              & $t_{rel}$                              & $r_{rel}$                              & $t_{rel}$                              & $r_{rel}$                              & $t_{rel}$                              & $r_{rel}$                              & $t_{rel}$                              & $r_{rel}$                              & $t_{rel}$                              & $r_{rel}$                              & $t_{rel}$                              & $r_{rel}$                              & $t_{rel}$                              & $r_{rel}$                               \\ 
\hhline{~--------------------}
VINS-Mono                          & {\cellcolor[rgb]{1,0.8,0.6}}2.08     & 1.68                                 & 4.27                                 & {\cellcolor[rgb]{1,0.8,0.6}}0.32     & 2.08                                 & {\cellcolor[rgb]{1,0.8,0.6}}0.63     & 3.22                                 & {\cellcolor[rgb]{1,0.8,0.6}}0.33     & 4.72                                 & 0.65                                 & {\cellcolor[rgb]{1,0.8,0.6}}1.89     & 0.17                                 & {\cellcolor[rgb]{1,0.8,0.6}}1.01     & {\cellcolor[rgb]{1,0.8,0.6}}0.20     & {\cellcolor[rgb]{1,0.8,0.6}}1.19     & 0.22                                 & {\cellcolor[rgb]{1,0.8,0.6}}1.35     & {\cellcolor[rgb]{1,0.8,0.6}}0.18     & {\cellcolor[rgb]{1,0.612,0.612}}3.61 & {\cellcolor[rgb]{1,0.8,0.6}}0.22      \\
ORB-SLAM3                          & 3.51                                 & 1.42                                 & 4.01                                 & 0.94                                 & 4.41                                 & 0.95                                 & 3.36                                 & 0.87                                 & 4.30                                 & 0.89                                 & 2.39                                 & {\cellcolor[rgb]{1,0.8,0.6}}0.12     & 1.31                                 & 0.22                                 & 1.41                                 & {\cellcolor[rgb]{1,0.612,0.612}}0.23 & 1.69                                 & 0.18                                 & 5.34                                 & {\cellcolor[rgb]{1,0.612,0.612}}0.21  \\
Selective-VIO                      & 2.41                                 & 0.78                                 & {\cellcolor[rgb]{1,0.612,0.612}}1.90 & 0.52                                 & {\cellcolor[rgb]{1,0.8,0.6}}1.72     & 1.01                                 & {\cellcolor[rgb]{1,0.8,0.6}}2.23     & 0.91                                 & {\cellcolor[rgb]{1,0.8,0.6}}2.83     & 0.80                                 & -                                    & -                                    & -                                    & -                                    & -                                    & -                                    & -                                    & -                                    & -                                    & -                                     \\
iSLAM                              & {\cellcolor[rgb]{1,0.612,0.612}}2.08 & {\cellcolor[rgb]{1,0.8,0.6}}0.53     & 2.40                                 & {\cellcolor[rgb]{1,0.612,0.612}}0.32 & 2.22                                 & {\cellcolor[rgb]{1,0.612,0.612}}0.47 & 2.78                                 & 0.43                                 & {\cellcolor[rgb]{1,0.612,0.612}}2.51 & {\cellcolor[rgb]{1,0.8,0.6}}0.41     & 7.75                                 & 0.36                                 & 38.46                                & 0.56                                 & 9.36                                 & 1.01                                 & 32.18                                & 1.46                                 & 4.74                                 & 0.36                                  \\
Ours                               & 2.64                                 & {\cellcolor[rgb]{1,0.612,0.612}}0.44 & {\cellcolor[rgb]{1,0.8,0.6}}2.01     & 0.40                                 & {\cellcolor[rgb]{1,0.612,0.612}}1.01 & 0.80                                 & {\cellcolor[rgb]{1,0.612,0.612}}1.90 & {\cellcolor[rgb]{1,0.612,0.612}}0.23 & 2.84                                 & {\cellcolor[rgb]{1,0.612,0.612}}0.38 & {\cellcolor[rgb]{1,0.612,0.612}}0.76 & {\cellcolor[rgb]{1,0.612,0.612}}0.10 & {\cellcolor[rgb]{1,0.612,0.612}}0.58 & {\cellcolor[rgb]{1,0.612,0.612}}0.17 & {\cellcolor[rgb]{1,0.612,0.612}}1.16 & {\cellcolor[rgb]{1,0.8,0.6}}0.23     & {\cellcolor[rgb]{1,0.612,0.612}}0.73 & {\cellcolor[rgb]{1,0.612,0.612}}0.16 & {\cellcolor[rgb]{1,0.8,0.6}}4.23     & 0.42                                  \\
\hline
\end{tabular}
\end{table*}

\subsection{Rendering Performance}
In this section, we compare the rendering performance of VINGS-Mono with both NeRF based methods and 3DGS based methods.

\subsubsection{Indoor Rendering Results}

We evaluated the rendering quality of our method in comparison to a NeRF-based approach GO-SLAM~\cite{GO-SLAM} and two state-of-the-art 3DGS-based monocular SLAM methods, MonoGS~\cite{MONO-GS} and PhotoSLAM~\cite{hhuang2024photoslam}, using the ScanNet~\cite{scannet} and BundleFusion~\cite{bundlefusion} datasets. For MonoGS, we initialized the process with 1000 iterations and performed 150 iterations for each subsequent frame. In the case of PhotoSLAM, due to its subpar performance under monocular settings (PSNR below 10), we used ~\cite{yin2023metric3d} predicted depth as a prior for the ScanNet dataset and recorded the rendered results, training for 100 iterations per image. Our method, designed for monocular settings, trained for just 80 iterations per image.

We present the qualitative and quantitative analysis of rendering quality for indoor scenes respectively. As illustrated in Figure~\ref{Fig:qualitative_rendering_results}, even though PhotoSLAM utilizes depth priors under RGB-D settings, our method still outperforms it across most scenes. During experiments, we observed that both PhotoSLAM and MonoGS frequently encounter issues where floaters cover the entire frame during later stages of tracking. In contrast, our method demonstrates greater stability, thanks to our management mechnism of gaussian map, enabling more robust and reliable performance. We used the poses estimated by each method as the viewpoints for image rendering and computed the average rendering quality for each scene over its trajectory sequence. Tab.~\ref{Tab:render_indoor} shows that our method achieves the best quantitative performance on both ScanNet and BundleFusion datasets, with the highest SSIM (0.79), lowest LPIPS (0.22 on ScanNet and 0.29 on BundleFusion), and best PSNR (22.43 dB on ScanNet and 20.97 dB on BundleFusion). These results validate the comprehensive superiority of our approach in terms of PSNR, LPIPS and SSIM.


\subsubsection{Outdoor Rendering Results}

Outdoor scenes present significantly greater challenges compared to indoor environments due to longer trajectory lengths (ranging from hundreds to thousands of meters) and faster movement speeds. These factors considerably impact the rendering quality and map storage requirements of baseline methods. We conducted experiments on five datasets: KITTI~\cite{KITTI}, KITTI-360~\cite{KITTI360}, Waymo~\cite{Waymo2021Tech}, Hierarchical~\cite{hierarchical_3dgs}, and MegaNeRF~\cite{meganerf}, which vary in lighting conditions, camera angles, and capturing devices. Due to the trajectories in KITTI and KITTI-360 often spanning several kilometers, no existing NeRF/3DGS-based SLAM method, apart from ours, has been successfully applied to them. To ensure fair comparisons, we trained the baseline methods on the first 500 frames, using the same settings as those for indoor experiments. Since the sky lacks depth and normal vectors, we used SegFormer to mask out the sky and computed the rendering quality metrics based on the masked results. Importantly, the same filtered datasets were used for all methods during evaluation to maintain fairness.

\begin{figure*}[!t]
\centering
\includegraphics[width=\textwidth]{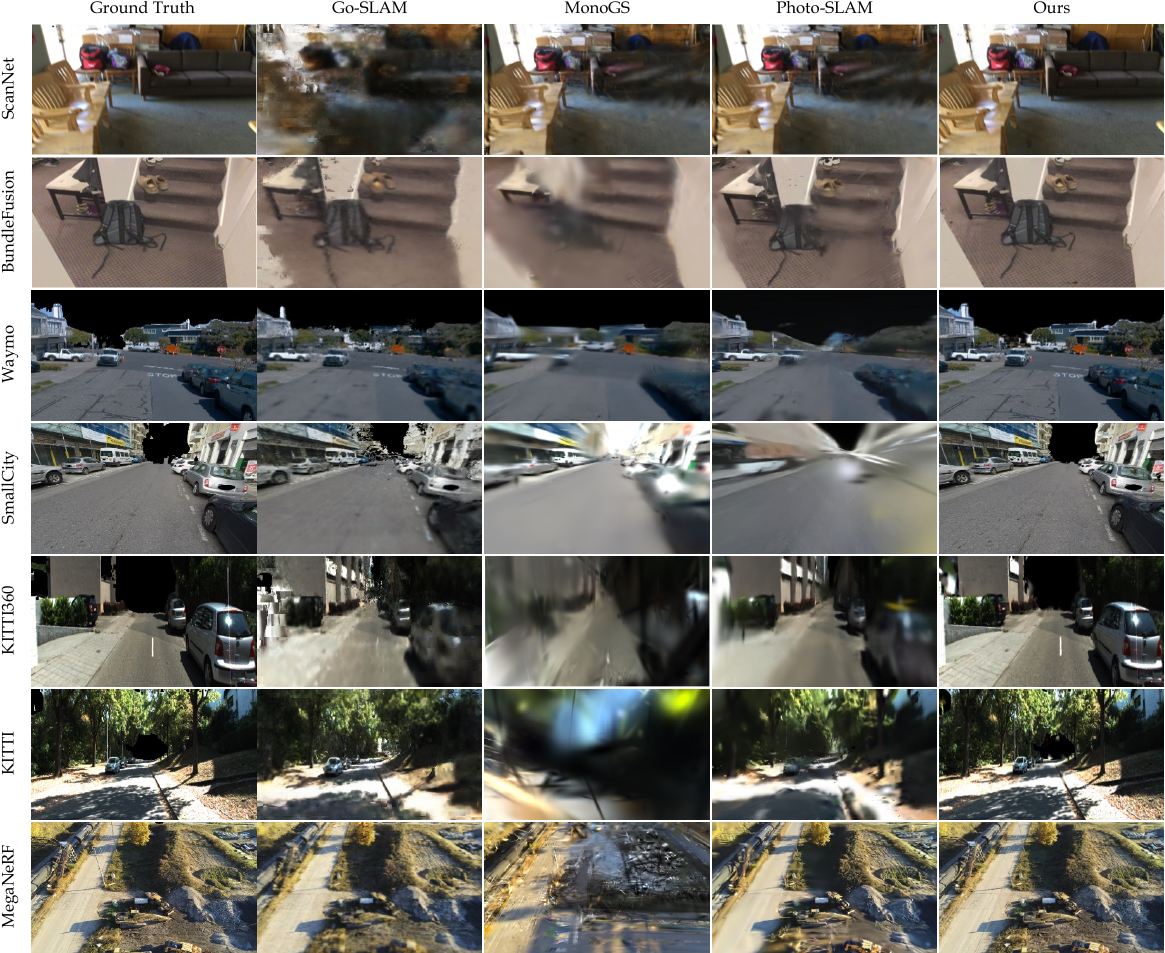}
\caption{\textbf{Qualitative Rendering Results}.We compared our method on two indoor~\cite{scannet, bundlefusion} and five outdoor scenes~\cite{Waymo2021Tech, hierarchical_3dgs, KITTI, KITTI360, meganerf}, with three advanced monocular SLAM algorithms, including the NeRF-based GO-SLAM~\cite{GO-SLAM} and two GS-based methods, MonoGS~\cite{MONO-GS} and PhotoSLAM~\cite{hhuang2024photoslam}. VINGS-Mono significantly outperforms existing methods in rendering quality. }
\label{Fig:qualitative_rendering_results}
\end{figure*}

As shown in Tab.~\ref{Fig:render_outdoor} and Fig.~\ref{Fig:qualitative_rendering_results}, our method excels in rendering high-quality details across large-scale environments and extended trajectories, even for long-distance, high-speed autonomous driving datasets such as KITTI, KITTI-360, and Waymo. For handheld datasets like Hierarchical, which are characterized by random motion trajectories and limited capture ranges, our approach achieves high-precision modeling of building edges and surface details. In drone datasets with significant depth variations and highly complex hierarchical scenes, our method demonstrates superior reconstruction quality, particularly in sparse regions. Across all scenarios, our system consistently achieves state-of-the-art (SOTA) performance in PSNR, SSIM, and LPIPS metrics, highlighting its robustness and versatility.

\begin{figure*}[!t]
\centering
\includegraphics[width=\textwidth]{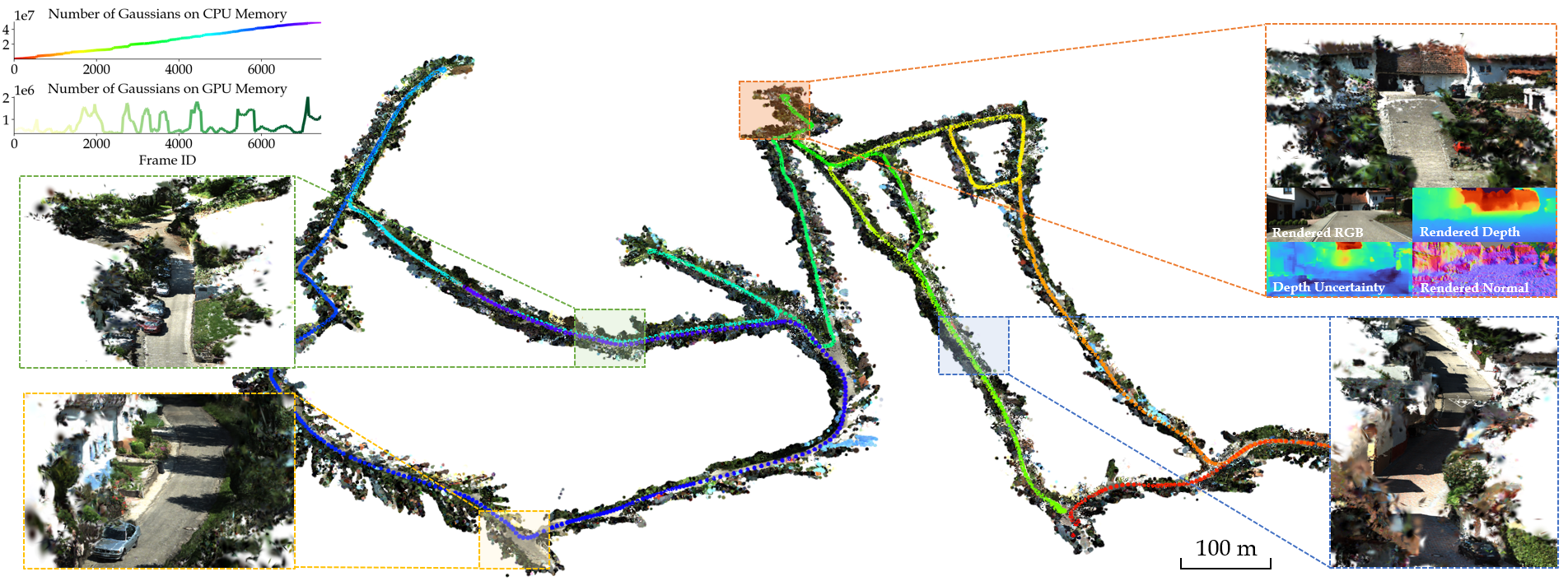}
\caption{\textbf{Visualization of KITTI360's gaussian map.} The trajectory length of scene 2013\_05\_28\_drive\_0006 is 8.05 km, and the entire Gaussian map contains 51.73 million ellipsoids. We recorded the number of Gaussians throughout the training process and zoomed in on different parts of the map for clearer visualization.}
\label{Fig:KITTI360Map}
\end{figure*}

We visualized the Gaussian map generated by our method on KITTI-360 from both a BEV (bird's eye view) and top-down view, we record the number of Gaussian ellipsoids on the GPU and CPU throughout the training process. As shown in Fig.~\ref{Fig:KITTI360Map}, our method robustly adapts to large-scale urban scenes. Unlike existing Gaussian SLAM methods, which struggle with inevitable floaters, our approach, supported by our score manager, produces clean and accurate geometry even in tree-dense regions, as illustrated by the green dashed boxes in Fig.~\ref{Fig:KITTI360Map}. Our method is capable of handling kilometer-scale scenes with 51.7 million Gaussians using a single RTX 4090 GPU.

\begin{figure*}[!t]
\centering
\includegraphics[width=\textwidth]{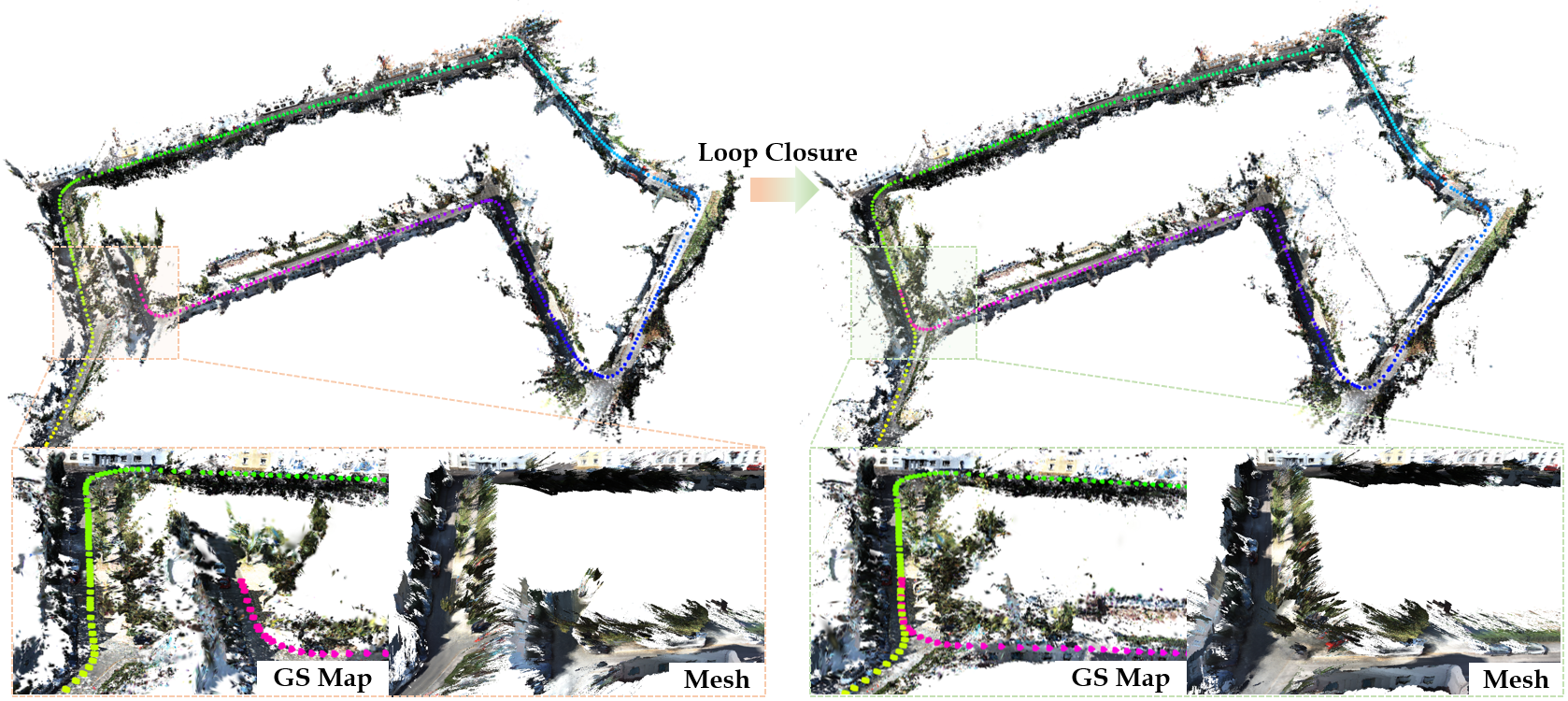}
\caption{\textbf{Performance of NVS Loop Closure in urban scenes.} Our NVS Loop Closure can correct the Gaussian map without time-consuming retraining. We zoomed in on the Gaussian map  at the loop closure location and export the mesh, our method effectively ensuring the global consistency of the gaussian map.}
\label{Fig:nvs_loopclosure}
\end{figure*}

In terms of map's global consistency, We demonstrated the impact of our loop closure on poses and the Gaussian map, as shown in Fig.~\ref{Fig:nvs_loopclosure}. For large-scale scenes, our method can directly correct the Gaussian map without retraining, ensuring the construction of a globally consistent map.

\begin{figure}[!t]
\centering
\includegraphics[width=0.5\textwidth]{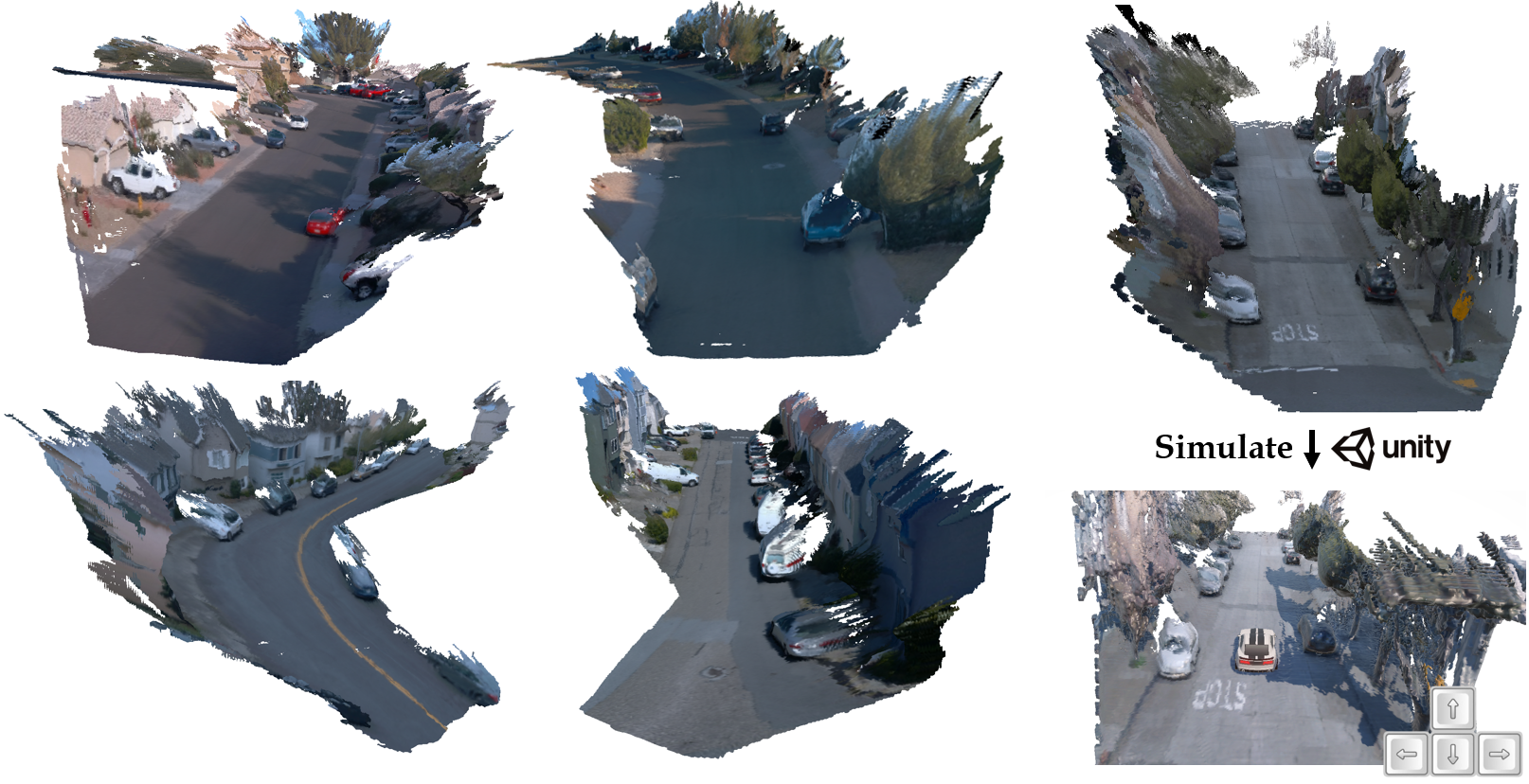}
\caption{\textbf{Mesh Extraction of Waymo Dataset.} We support extracting meshes based on rendered depths and loading them into Unity for simulation to narrow the gap of simulation and reality .}
\label{Fig:waymo_mesh}
\end{figure}

Additionally, our method supports rendering color and depth from interpolated poses and exporting a mesh using TSDF-Fusion~\cite{tsdf-fusion}. This further expands the application scope of our approach. As shown in Fig.~\ref{Fig:waymo_mesh}, we exported a mesh for the Waymo dataset and performed simulation in Unity.



\subsection{Dynamic Eraser Performance}

To evaluate the effectiveness of Dynamic Eraser in enhancing tracking performance in dynamic environments, we masked out dynamic objects in the frontend BA using Dynamic Eraser. We selected the dynamic SLAM dataset, BONN Dataset~\cite{bonn_dataset}, and conducted comparative experiments with two SLAM methods tailored for dynamic scenes, as shown in Tab.~\ref{Tab:dynamic_localiztion}. Note that, since the baseline methods require depth information, all results presented in the table are based on RGB-D input for Dynamic SLAM. Compared with ReFusion~\cite{refusion} and RodynSLAM~\cite{rodyn_slam}, our approach achieves superior results. Additionally, we performed ablation studies, and the results demonstrate that for datasets where dynamic objects dominate the scene, Dynamic Eraser significantly improves tracking accuracy.

\begin{table}
\centering
\caption{Localization results on several dynamic scene sequences in the BONN dataset~\cite{bonn_dataset}.}
\label{Tab:dynamic_localiztion}
\begin{tblr}{
  cells = {c},
  vline{2} = {-}{},
  hline{1-2,6} = {-}{},
}
ATE [cm] ↓        & ball  & ps\_tk & ps\_tk2 & mv\_box2 & Avg.  \\
ReFusion         & 17.5  & 28.9   & 46.3    & 17.9     & 27.65 \\
RodynSLAM        & 7.9   & 14.5   & 13.8    & 12.6     & 12.2  \\
Ours (wo Eraser) & 11.75 & 37.48  & 48.31   & 23.44    & 30.25 \\
Ours (w Eraser)  & 4.08  & 4.63   & 5.05    & 3.58     & 4.34  
\end{tblr}
\end{table}

\begin{table*}
\centering
\caption{Quantitative results on the indoor datasets Scannet and BundleFusion. We mark the best two results with \colorbox{red}{first} and \colorbox{orange}{second}. All quantitative metrics are computed as averages based on renderings at the same keyframes.}
\label{Tab:render_indoor}
\begin{tabular}{c|c|cccccc|ccccc} 
\hline
\multicolumn{1}{c}{}       &        & \multicolumn{6}{c|}{ScanNet}                                                                                                                                                                                          & \multicolumn{5}{c}{BundleFusion}                                                                                                                                                   \\
\multicolumn{1}{c}{}       &        & 0054                              & 0059                              & 0106                              & 0169                              & 0233                              & 0465                              & apt0                              & apt2                              & copyroom                          & office0                           & office2                            \\ 
\hline
\multirow{3}{*}{GO-SLAM}   & SSIM↑  & 0.59                              & 0.32                              & 0.47                              & 0.42                              & 0.48                              & 0.09                              & 0.52                              & 0.34                              & 0.61                              & 0.23                              & 0.51                               \\
                           & LPIPS↓ & 0.53                              & 0.60                              & 0.59                              & 0.57                              & 0.55                              & 0.75                              & {\cellcolor[rgb]{1,0.8,0.6}}0.54  & {\cellcolor[rgb]{1,0.8,0.6}}0.59  & 0.49                              & 0.72                              & {\cellcolor[rgb]{1,0.8,0.6}}0.55   \\
                           & PSNR↑  & 19.70                             & 13.15                             & 14.58                             & 14.49                             & 17.22                             & 8.65                              & 17.24                             & {\cellcolor[rgb]{1,0.8,0.6}}12.24 & {\cellcolor[rgb]{1,0.8,0.6}}18.40 & 12.60                             & 17.31                              \\ 
\hline
\multirow{3}{*}{MonoGS}    & SSIM↑  & 0.83                              & 0.74                              & 0.76                              & 0.78                              & 0.74                              & 0.69                              & {\cellcolor[rgb]{1,0.8,0.6}}0.74  & 0.39                              & {\cellcolor[rgb]{1,0.6,0.6}}0.78  & {\cellcolor[rgb]{1,0.6,0.6}}0.68  & {\cellcolor[rgb]{1,0.8,0.6}}0.67   \\
                           & LPIPS↓ & 0.61                              & 0.59                              & 0.60                              & 0.61                              & 0.67                              & 0.74                              & 0.62                              & 0.82                              & 0.57                              & 0.68                              & 0.67                               \\
                           & PSNR↑  & {\cellcolor[rgb]{1,0.8,0.6}}21.37 & {\cellcolor[rgb]{1,0.8,0.6}}18.55 & {\cellcolor[rgb]{1,0.8,0.6}}17.58 & {\cellcolor[rgb]{1,0.8,0.6}}19.15 & 19.73                             & 17.19                             & {\cellcolor[rgb]{1,0.8,0.6}}18.80 & 11.50                             & 17.83                             & {\cellcolor[rgb]{1,0.8,0.6}}16.76 & {\cellcolor[rgb]{1,0.8,0.6}}18.98  \\ 
\hline
\multirow{3}{*}{PhotoSLAM} & SSIM↑  & {\cellcolor[rgb]{1,0.8,0.6}}0.83  & {\cellcolor[rgb]{1,0.8,0.6}}0.772 & {\cellcolor[rgb]{1,0.8,0.6}}0.78  & {\cellcolor[rgb]{1,0.8,0.6}}0.79  & {\cellcolor[rgb]{1,0.6,0.6}}0.78  & {\cellcolor[rgb]{1,0.6,0.6}}0.74  & 0.66                              & {\cellcolor[rgb]{1,0.8,0.6}}0.59  & 0.73                              & 0.43                              & 0.33                               \\
                           & LPIPS↓ & {\cellcolor[rgb]{1,0.8,0.6}}0.35  & {\cellcolor[rgb]{1,0.8,0.6}}0.41  & {\cellcolor[rgb]{1,0.8,0.6}}0.37  & {\cellcolor[rgb]{1,0.8,0.6}}0.39  & {\cellcolor[rgb]{1,0.8,0.6}}0.37  & {\cellcolor[rgb]{1,0.8,0.6}}0.45  & 0.56                              & 0.60                              & {\cellcolor[rgb]{1,0.8,0.6}}0.36  & {\cellcolor[rgb]{1,0.8,0.6}}0.63  & 0.68                               \\
                           & PSNR↑  & 20.54                             & 17.17                             & 16.09                             & 17.46                             & {\cellcolor[rgb]{1,0.6,0.6}}23.95 & {\cellcolor[rgb]{1,0.8,0.6}}19.88 & 11.46                             & 11.68                             & 16.96                             & 9.21                              & 8.55                               \\ 
\hline
\multirow{3}{*}{Ours}      & SSIM↑  & {\cellcolor[rgb]{1,0.6,0.6}}0.84  & {\cellcolor[rgb]{1,0.6,0.6}}0.775 & {\cellcolor[rgb]{1,0.6,0.6}}0.83  & {\cellcolor[rgb]{1,0.6,0.6}}0.80  & {\cellcolor[rgb]{1,0.8,0.6}}0.77  & {\cellcolor[rgb]{1,0.8,0.6}}0.69  & {\cellcolor[rgb]{1,0.6,0.6}}0.75  & {\cellcolor[rgb]{1,0.6,0.6}}0.63  & {\cellcolor[rgb]{1,0.8,0.6}}0.74  & {\cellcolor[rgb]{1,0.8,0.6}}0.65  & {\cellcolor[rgb]{1,0.6,0.6}}0.68   \\
                           & LPIPS↓ & {\cellcolor[rgb]{1,0.6,0.6}}0.20  & {\cellcolor[rgb]{1,0.6,0.6}}0.24  & {\cellcolor[rgb]{1,0.6,0.6}}0.18  & {\cellcolor[rgb]{1,0.6,0.6}}0.22  & {\cellcolor[rgb]{1,0.6,0.6}}0.22  & {\cellcolor[rgb]{1,0.6,0.6}}0.25  & {\cellcolor[rgb]{1,0.6,0.6}}0.28  & {\cellcolor[rgb]{1,0.6,0.6}}0.41  & {\cellcolor[rgb]{1,0.6,0.6}}0.33  & {\cellcolor[rgb]{1,0.6,0.6}}0.39  & {\cellcolor[rgb]{1,0.6,0.6}}0.23   \\
                           & PSNR↑  & {\cellcolor[rgb]{1,0.6,0.6}}26.31 & {\cellcolor[rgb]{1,0.6,0.6}}20.51 & {\cellcolor[rgb]{1,0.6,0.6}}23.10 & {\cellcolor[rgb]{1,0.6,0.6}}22.27 & {\cellcolor[rgb]{1,0.8,0.6}}23.67 & {\cellcolor[rgb]{1,0.6,0.6}}21.27 & {\cellcolor[rgb]{1,0.6,0.6}}20.45 & {\cellcolor[rgb]{1,0.6,0.6}}18.61 & {\cellcolor[rgb]{1,0.6,0.6}}18.47 & {\cellcolor[rgb]{1,0.6,0.6}}19.85 & {\cellcolor[rgb]{1,0.6,0.6}}22.23  \\
\hline
\end{tabular}
\end{table*}

\begin{table*}
\centering

\caption{Quantitative analysis results on the outdoor datasets KITTI, KITTI360, Waymo, Hierarchical, and MegaNeRF. ``-'' indicates that the system failed to track and render images in the whole scenario. }
\label{Fig:render_outdoor}
\begin{tabular}{c|c|ccc|ccc|ccc|cc|cc} 
\hline
\multicolumn{1}{c}{}       &        & \multicolumn{3}{c|}{KITTI}                                                                                & \multicolumn{3}{c|}{KITTI360}                                                                             & \multicolumn{3}{c|}{Waymo}                                                                                & \multicolumn{2}{c|}{Hierarchical}                                     & \multicolumn{2}{c}{MegaNeRF}                                           \\
\multicolumn{1}{c}{}       &        & 02                                & 07                                & 08                                & 05                                & 06                                & 10                                & 01                                & 03                                & 14                                & SmallCity                         & Campus                            & Building                          & Rubble                             \\ 
\hline
\multirow{3}{*}{GO-SLAM}   & SSIM↑  & 0.39                              & 0.46                              & 0.51                              & 0.44                              & 0.43                              & 0.38                              & 0.78                              & 0.70                              & 0.63                              & 0.33                              & 0.33                              & {\cellcolor[rgb]{1,0.8,0.6}}0.53  & {\cellcolor[rgb]{1,0.8,0.6}}0.63   \\
                           & LPIPS↓ & {\cellcolor[rgb]{1,0.8,0.6}}0.49  & {\cellcolor[rgb]{1,0.8,0.6}}0.45  & {\cellcolor[rgb]{1,0.8,0.6}}0.43  & {\cellcolor[rgb]{1,0.8,0.6}}0.47  & {\cellcolor[rgb]{1,0.8,0.6}}0.45  & {\cellcolor[rgb]{1,0.8,0.6}}0.20  & {\cellcolor[rgb]{1,0.8,0.6}}0.20  & {\cellcolor[rgb]{1,0.8,0.6}}0.30  & {\cellcolor[rgb]{1,0.8,0.6}}0.34  & {\cellcolor[rgb]{1,0.8,0.6}}0.57  & {\cellcolor[rgb]{1,0.8,0.6}}0.54  & {\cellcolor[rgb]{1,0.8,0.6}}0.40  & {\cellcolor[rgb]{1,0.8,0.6}}0.32   \\
                           & PSNR↑  & 15.01                             & 12.81                             & 14.62                             & 14.27                             & 14.24                             & {\cellcolor[rgb]{1,0.8,0.6}}21.07 & 21.07                             & {\cellcolor[rgb]{1,0.8,0.6}}21.22 & 19.54                             & {\cellcolor[rgb]{1,0.8,0.6}}14.30 & 13.41                             & {\cellcolor[rgb]{1,0.8,0.6}}20.71 & {\cellcolor[rgb]{1,0.8,0.6}}20.81  \\ 
\hline
\multirow{3}{*}{MonoGS}    & SSIM↑  & 0.34                              & 0.43                              & {\cellcolor[rgb]{1,0.8,0.6}}0.52  & {\cellcolor[rgb]{1,0.8,0.6}}0.53  & 0.55                              & 0.20                              & {\cellcolor[rgb]{1,0.8,0.6}}0.83  & {\cellcolor[rgb]{1,0.8,0.6}}0.74  & {\cellcolor[rgb]{1,0.8,0.6}}0.82  & -                                 & 0.52                              & 0.23                              & 0.24                               \\
                           & LPIPS↓ & 0.85                              & 0.78                              & 0.75                              & 0.68                              & 0.61                              & 0.85                              & 0.40                              & 0.63                              & 0.56                              & -                                 & 0.72                              & 0.96                              & 0.94                               \\
                           & PSNR↑  & 10.63                             & 12.59                             & {\cellcolor[rgb]{1,0.8,0.6}}15.01 & {\cellcolor[rgb]{1,0.8,0.6}}16.08 & 15.63                             & 10.20                             & {\cellcolor[rgb]{1,0.8,0.6}}22.63 & 19.29                             & {\cellcolor[rgb]{1,0.8,0.6}}23.00 & -                                 & {\cellcolor[rgb]{1,0.8,0.6}}14.49 & 11.06                             & 11.50                              \\ 
\hline
\multirow{3}{*}{PhotoSLAM} & SSIM↑  & {\cellcolor[rgb]{1,0.8,0.6}}0.44  & {\cellcolor[rgb]{1,0.8,0.6}}0.52  & 0.48                              & 0.51                              & {\cellcolor[rgb]{1,0.8,0.6}}0.56  & {\cellcolor[rgb]{1,0.8,0.6}}0.51  & 0.74                              & 0.69                              & 0.76                              & {\cellcolor[rgb]{1,0.8,0.6}}0.39  & {\cellcolor[rgb]{1,0.8,0.6}}0.57  & 0.31                              & 0.27                               \\
                           & LPIPS↓ & 0.66                              & 0.56                              & 0.65                              & 0.55                              & 0.49                              & 0.65                              & 0.39                              & 0.47                              & 0.42                              & 0.71                              & 0.56                              & 0.76                              & 0.67                               \\
                           & PSNR↑  & {\cellcolor[rgb]{1,0.8,0.6}}15.25 & {\cellcolor[rgb]{1,0.8,0.6}}15.03 & 14.25                             & 15.57                             & {\cellcolor[rgb]{1,0.8,0.6}}15.81 & 14.78                             & 15.08                             & 15.35                             & 15.99                             & 11.57                             & 11.40                             & 15.47                             & 14.09                              \\ 
\hline
\multirow{3}{*}{Ours}      & SSIM↑  & {\cellcolor[rgb]{1,0.6,0.6}}0.68  & {\cellcolor[rgb]{1,0.6,0.6}}0.73  & {\cellcolor[rgb]{1,0.6,0.6}}0.79  & {\cellcolor[rgb]{1,0.6,0.6}}0.80  & {\cellcolor[rgb]{1,0.6,0.6}}0.80  & {\cellcolor[rgb]{1,0.6,0.6}}0.82  & {\cellcolor[rgb]{1,0.6,0.6}}0.85  & {\cellcolor[rgb]{1,0.6,0.6}}0.86  & {\cellcolor[rgb]{1,0.6,0.6}}0.85  & {\cellcolor[rgb]{1,0.6,0.6}}0.81  & {\cellcolor[rgb]{1,0.6,0.6}}0.78  & {\cellcolor[rgb]{1,0.6,0.6}}0.82  & {\cellcolor[rgb]{1,0.6,0.6}}0.82   \\
                           & LPIPS↓ & {\cellcolor[rgb]{1,0.6,0.6}}0.26  & {\cellcolor[rgb]{1,0.6,0.6}}0.29  & {\cellcolor[rgb]{1,0.6,0.6}}0.27  & {\cellcolor[rgb]{1,0.6,0.6}}0.17  & {\cellcolor[rgb]{1,0.6,0.6}}0.17  & {\cellcolor[rgb]{1,0.6,0.6}}0.16  & {\cellcolor[rgb]{1,0.6,0.6}}0.18  & {\cellcolor[rgb]{1,0.6,0.6}}0.16  & {\cellcolor[rgb]{1,0.6,0.6}}0.19  & {\cellcolor[rgb]{1,0.6,0.6}}0.22  & {\cellcolor[rgb]{1,0.6,0.6}}0.21  & {\cellcolor[rgb]{1,0.6,0.6}}0.15  & {\cellcolor[rgb]{1,0.6,0.6}}0.15   \\
                           & PSNR↑  & {\cellcolor[rgb]{1,0.6,0.6}}19.96 & {\cellcolor[rgb]{1,0.6,0.6}}20.15 & {\cellcolor[rgb]{1,0.6,0.6}}20.93 & {\cellcolor[rgb]{1,0.6,0.6}}24.52 & {\cellcolor[rgb]{1,0.6,0.6}}22.82 & {\cellcolor[rgb]{1,0.6,0.6}}24.47 & {\cellcolor[rgb]{1,0.6,0.6}}23.48 & {\cellcolor[rgb]{1,0.6,0.6}}24.72 & {\cellcolor[rgb]{1,0.6,0.6}}23.76 & {\cellcolor[rgb]{1,0.6,0.6}}22.07 & {\cellcolor[rgb]{1,0.6,0.6}}21.46 & {\cellcolor[rgb]{1,0.6,0.6}}25.45 & {\cellcolor[rgb]{1,0.6,0.6}}25.21  \\
\hline
\end{tabular}
\end{table*}

\subsection{Ablation Studies}

\subsubsection{Ablation on Sample Rasterizer}\label{subsubsec:sample_rasterizer}

\begin{figure}[!t]
\centering
\includegraphics[width=0.5\textwidth]{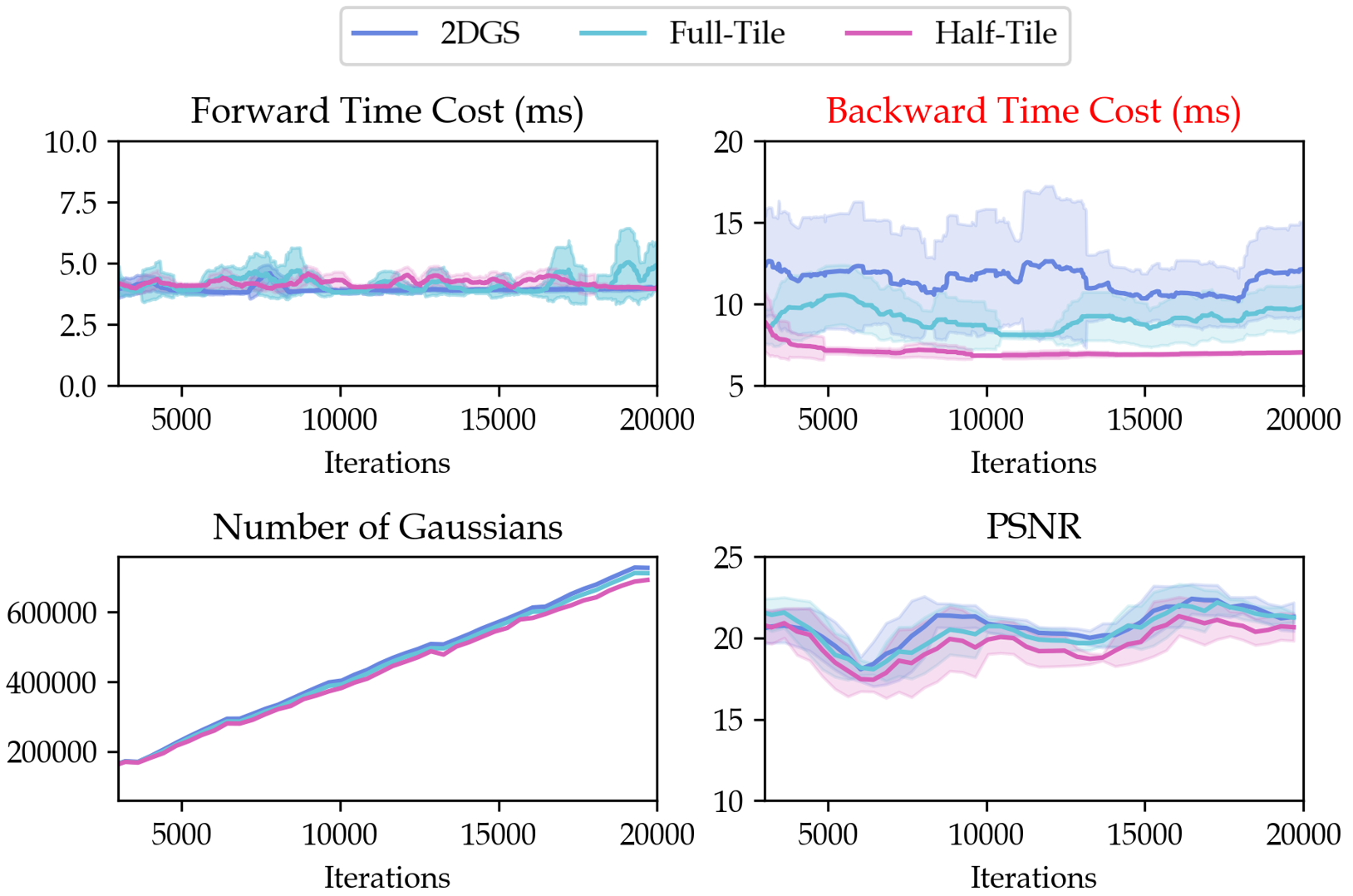}
\caption{Ablation on sample rasterizer.}
\label{Fig:ablation_on_sample_rasterizer}
\end{figure}

We selected KITTI dataset to profile the rendering during forward propagation and backpropagation for each training iteration. As shown in Fig.~\ref{Fig:ablation_on_sample_rasterizer}, we recorded the number of Gaussians, PSNR values, and the time taken for forward and backward propagation. We compared our approach with the original 2DGS pixel-parallel method and the Gaussian-parallel method used in Taming3DGS. Since our acceleration strategy does not affect the score manager, the number of Gaussians showed no significant change. In terms of training speed, compared to performing backpropagation on all pixels within a tile (Full-Tile), our sample-based backpropagation strategy achieved a 170\% acceleration in backpropagation time (from 7.21 ms to 4.23 ms), despite a 0.56 drop in average PSNR. Compared to the original 2DGS approach, our method accelerated backpropagation by 273\% (from 11.55 ms to 4.23 ms).

\subsubsection{Ablation on Score Manager}

The primary function of our score manager is to remove unnecessary Gaussians and facilitate data transfer between the CPU and GPU. This is achieved by reducing the number of Gaussians while minimizing any impact on rendering quality. To evaluate its effectiveness, we conducted ablation studies on one indoor and one outdoor scenes. As shown in Tab. ~\ref{Tab:abalation_on_score_manager}, our score manager significantly reduces the number of Gaussians. Surprisingly, for indoor scene, it not only reduces the Gaussian count but also improves rendering quality. For outdoor scene, it achieves a 22\% reduction in Gaussian ellipsoids while maintaining a PSNR decrease of only 0.3\%.

\begin{table}

\centering
\caption{Ablation on Score Manager.}
\label{Tab:abalation_on_score_manager}
\begin{tblr}{
  cells = {c},
  cell{1}{2} = {c=2}{},
  cell{1}{4} = {c=2}{},
  vline{2-3,5} = {1}{},
  vline{2,4} = {2-7}{},
  hline{1,3,8} = {-}{},
  hline{2} = {2-5}{},
}
         & ScanNet-0106   &           & Waymo-Scene13     &           \\
         & Avg. PSNR & GS Number & Avg. PSNR & GS Number \\
0.0 (wo) & 22.98     & 4,041,325 & 23.67     & 1,777,807 \\
0.8      & 22.58     & 3,104,080 & 23.60     & 1,376,648 \\
12.8     & 23.07     & 2,675,419 & 23.47     & 1,321,745 \\
25.6     & 23.13     & 2,265,721 & 23.16     & 1,308,828 \\
102.4    & 23.02     & 1,964,771 & 22.47     & 1,059,930 
\end{tblr}
\end{table}

\begin{table}
\centering
\caption{Ablation on Pose Refinement.}
\label{Tab:abalation_on_pose_refinement}
\begin{tblr}{
  cells = {c},
  vline{2} = {-}{},
  hline{1-2,5} = {-}{},
}
ATE [m]↓                     & ScanNet-0106 & Copyroom & Campus \\
wo pose refine              & 0.25         & 0.83             & 1.83        \\
w refine current pose (~\cite{MONO-GS,SplaTAM})     & 0.19         & 0.64             & 1.19        \\
w refine visible poses (ours) & 0.16         & 0.39             & 1.03        
\end{tblr}
\end{table}

\subsubsection{Ablation on Pose Refinement}

Our method binds Gaussians to different keyframe poses, the proposed pose refinement strategy enables rendering a frame while simultaneously optimizing the poses of all visible keyframes. Existing GS-based SLAM~\cite{MONO-GS,SplaTAM} methods typically optimize the pose of the current frame using rerendering losses. To validate the effectiveness of our pose refinement strategy, we conducted ablation studies, as shown in Tab.~\ref{Tab:abalation_on_pose_refinement}. Our method outperforms the single-frame optimization strategy in both indoor and outdoor scenes. As shown in the Tab~\ref{Tab:abalation_on_pose_refinement}, our approach achieves significant improvements in scenes where the frontend tracking performance was initially poor, consistently outperforming existing single-frame optimization methods.

\subsection{Runtime Analysis}
We report the runtime and model size on three datasets with varying frame counts: Waymo, Hierarchical, and KITTI. Our VIO Frontend and Mapping modules run as two parallel threads, with the mapping speed being slower than tracking. First, we independently tested the runtime of the tracking module. Then, we disabled the visualization of the BEV (bird’s-eye-view) map to measure the overall runtime of our framework. The Model Size refers to the file size of the final Gaussian point cloud that is saved, which differs from the GPU memory usage during runtime. All results were profiled using an RTX 4090 GPU, as shown in Tab.~\ref{Tab:runtime_analysis}. Our method demonstrates the capability of running online for both shorter trajectories (e.g., around 300 meters, as in the Waymo dataset) and longer trajectories (3.2km, as in the KITTI dataset).

\subsection{Real world Experiments}

\begin{figure*}[!t]
\centering
\includegraphics[width=\textwidth]{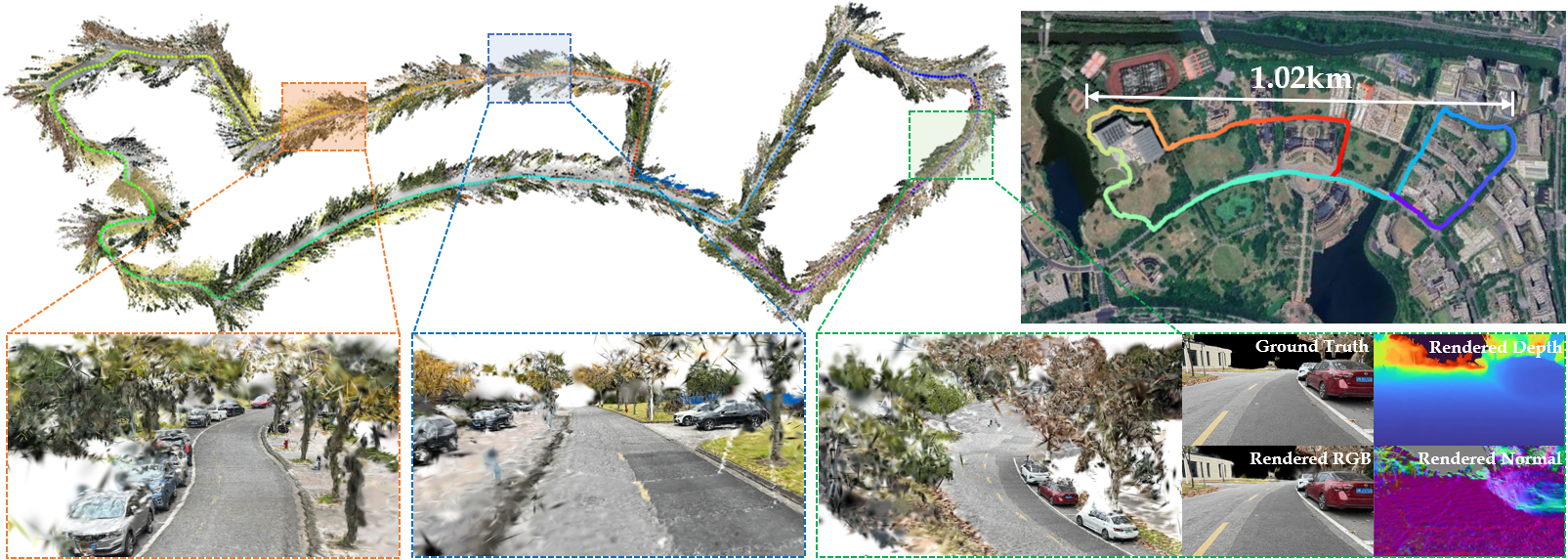}
\caption{\textbf{Monocular SLAM results of large scale self-collected data.} The collected data covers our campus. The trajectory and map on the left represent the results estimated by VINGS-Mono, while the top-right shows the smartphone GPS data recorded during data collection, aligned with Google Map.}
\label{Fig:jiangwan-campus}
\end{figure*}

\begin{figure}[!t]
\centering
\includegraphics[width=0.5\textwidth]{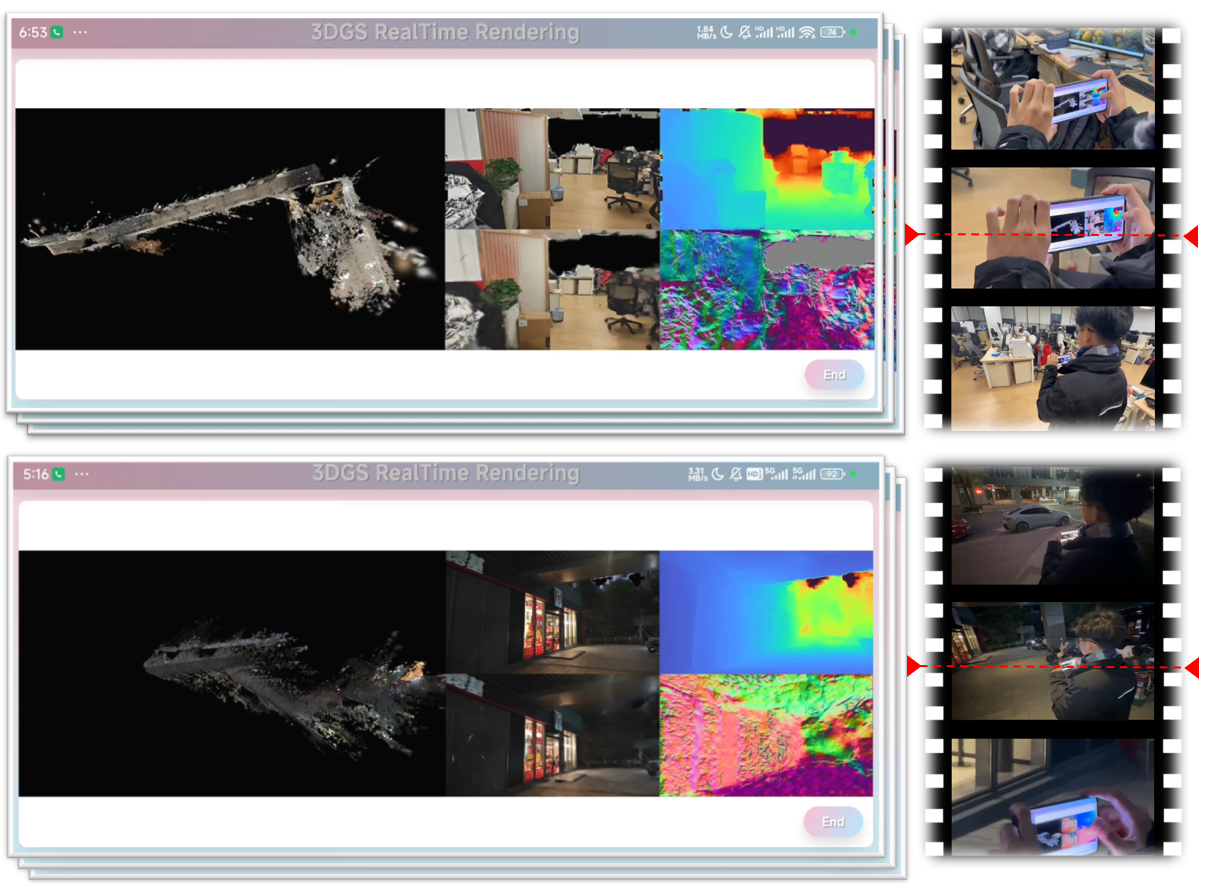}
\caption{\textbf{Mobile App of VINGS-Mono. }}
\label{Fig:mobileapp}
\end{figure}

\subsubsection{Large-scale Environment} 
To test the stability and robustness of our method, we collected a large outdoor dataset. This dataset covers our entire Campus and was recorded using an iPhone device. It spans approximately 1.02 km in length and 0.4 km in width. The data collection was conducted on a bike, riding at a speed of around 10 km/h. The dataset contains only 20Hz RGB image data (1280$\times$720) and 1Hz GPS data. We validated the performance of our algorithm in large-scale scenarios under monocular settings. As shown in Fig.~\ref{Fig:jiangwan-campus}, we visualized the GPS data in the top-right corner of the figure. It can be observed that even with only monocular input data, our method exhibited almost no scale drift in large-scale scenarios during long-duration tests.

\subsubsection{Mobile App on Smartphone}
We ran VINGS-Mono in a mobile phone setup. We developed an app using the Flutter framework, which can be deployed on iOS, Android, and Windows platforms. The app collects images with a resolution of 480×720 at 30 Hz, along with IMU data (from the built-in gyroscope and accelerometer). This data is then transferred to our server. On the server, VINGS-Mono processes the data and generates real-time visual outputs displayed on the phone screen. These outputs include a bird's-eye view Gaussian map, the captured images, rendered images, rendered depth maps, and rendered normal maps. To achieve better geometry in real-world scenarios, we utilize ~\cite{yin2023metric3d} to provide depth priors. It is important to note that this depth information is noisy and exhibits scale inconsistencies across multiple views. To evaluate the robustness of VINGS-Mono, we conducted experiments in both indoor and outdoor scenes under various lighting conditions. The experimenter held the phone and walked through our lab and around the square.
As shown in Fig.~\ref{Fig:mobileapp}, our method demonstrates strong performance in reconstructing low-texture regions such as white walls. Additionally, in outdoor scenes under low-light conditions, our method achieves high-quality reconstruction of highly exposed areas, such as illuminated signboards.

\begin{table}
\centering
\caption{Runtime Analysis.}
\label{Tab:runtime_analysis}
\begin{tblr}{
  colspec = {X[2.5] X[1] X[1] X[1] X[0.6] X[1.7]},
  cells = {c},
  cell{1}{1} = {r=2}{},
  cell{1}{5} = {r=2}{},
  cell{1}{6} = {r=2}{},
  vline{2} = {1-2,3-5}{},
  hline{1,3,6} = {-}{},
}
Dataset-Scene    & Frame  & Total   & Tracking & FPS  & Model Size \\
                 & Number & Runtime & /Frame     &      &            \\
Waymo-Scene01    & 198    & 117s    & 214ms     & 1.69 & 386Mb      \\
Hiera.-SmallCity & 877    & 739s    & 247ms     & 1.18 & 1817Mb     \\
KITTI-Odom08     & 5177   & 4560s   & 273ms     & 1.13 & 10366Mb    \\
\end{tblr}
\end{table}

\section{Conclusions and Future Work}

\subsection{Conclusions}
In this paper, we proposed VINGS-Mono, a monocular (inertial) Gaussian Splatting SLAM framework designed to address the challenges of large-scale environments. By introducing innovations such as a score manager for efficient Gaussian pruning, a single-to-multi pose refinement module to enhance tracking accuracy, a loop closure method leveraging novel view synthesis for global consistence, and a dynamic object masking mechanism to handle transient objects, VINGS-Mono achieves efficient, scalable, and accurate SLAM performance.

Our system was rigorously evaluated through extensive experiments. First, we conducted comparative experiments on two public indoor datasets and five outdoor datasets to assess the localization accuracy and rendering quality of VINGS-Mono. Comparisons with state-of-the-art NeRF/GS-based methods and visual SLAM approaches demonstrate the superior localization and mapping performance of our system. Additionally, we carried out real-world experiments in large-scale environments to validate the robustness and stability of our method. Next, we performed ablation studies on the individual modules of VINGS-Mono to verify their effectiveness. Finally, we developed a mobile application and validated the system's real-time capabilities through live demonstrations, showcasing the construction process of the 2D Gaussian map in both indoor and outdoor environments.

Our approach enables the creation of denser, high-quality maps by leveraging Gaussian Splatting, which reconstructs dense geometric and color information even in outdoor scenarios where LiDAR or depth cameras are impractical. This facilitates efficient navigation and exploration by preserving critical scene details and enabling advanced tasks like instance image-goal navigation. Additionally, Gaussian maps with novel view rendering capabilities are ideal for real-time applications in VR/AR and digital twins, enhancing scalability, adaptability, and efficiency for large-scale autonomous systems.

\subsection{Limitations and Future Works}
A key limitation of our work lies in its inability to effectively reconstruct and localize under extremely high-speed motion. Specifically, DBA faces challenges in capturing and recovering dense geometric information when frame intervals are large, while the multiple training iterations required by GS limit the reconstruction speed of the 2D Gaussian map. To address this, our future work will focus on integrating additional priors into DBA and incorporating networks such as ~\cite{pointtransformer,lsm} to directly output Gaussian attributes, thereby reducing the number of training iterations. Another limitation is that our system has not yet been deployed for on-device computation. In future work, we will explore deploying VINGS-Mono directly onto edge computing devices~\cite{mobile_3dgs} to further enhance the practical value and applicability of our algorithm.


\bibliographystyle{elsarticle-num} 
\bibliography{egbib}{}



\end{document}